\pdfoutput=1
\documentclass{article}
\usepackage{iclr2027_conference,times}

\usepackage{amsmath,amsfonts,bm}

\def\eqref#1{equation~\ref{#1}}

\def\plaineqref#1{\ref{#1}}

\def\1{\bm{1}}

\DeclareMathAlphabet{\mathsfit}{\encodingdefault}{\sfdefault}{m}{sl}
\SetMathAlphabet{\mathsfit}{bold}{\encodingdefault}{\sfdefault}{bx}{n}

\usepackage[utf8]{inputenc}
\usepackage[T1]{fontenc}
\usepackage[table]{xcolor}
\usepackage{booktabs}
\usepackage{amsfonts}
\usepackage{amsmath}
\usepackage{amssymb}
\usepackage{nicefrac}
\usepackage{graphicx}
\usepackage{subcaption}
\usepackage{tikz}
\usetikzlibrary{arrows.meta, positioning, shapes.geometric, fit, backgrounds, calc}
\usepackage{multirow}
\usepackage{amsthm}

\usepackage{algorithm}
\usepackage{algpseudocode}
\usepackage{enumitem}

\usepackage{hyperref}
\hypersetup{colorlinks=true, linkcolor=red!60!black, citecolor=blue!60!black, urlcolor=blue!80!black}
\usepackage{url}

\newcommand{\steamrow}{\rowcolor{gray!12}}

\algnewcommand\algorithmicinput{\textbf{Input:}}
\algnewcommand\algorithmicoutput{\textbf{Output:}}
\algnewcommand\Input{\item[\algorithmicinput]}
\algnewcommand\Output{\item[\algorithmicoutput]}

\newcommand{\Dexp}{\mathcal{D}_{\mathrm{exp}}}
\newcommand{\Dnexp}{\mathcal{D}_{\mathrm{nexp}}}

\newcommand{\Asteam}{A_{\mathrm{STEAM}}}

\usepackage{placeins}
\title{STEAM: Self-Supervised Temporal Ensemble Advantage Modeling for Real-World Robot Learning}

\author{\normalfont
  \begin{minipage}{\textwidth}
    \centering
    \normalsize
    \mbox{\textbf{Zhihao Liu}\textsuperscript{1,4,5,*}},
    \mbox{\textbf{Qiuyi Gu}\textsuperscript{2,3,6,*}},
    \mbox{\textbf{Yitao Wang}\textsuperscript{2}},
    \mbox{\textbf{Dongming Qiao}\textsuperscript{2}},
    \mbox{\textbf{Yixian Zhang}\textsuperscript{2}},
    \mbox{\textbf{Shuaihang Chen}\textsuperscript{7,5}},
    \mbox{\textbf{Liangzhi Shi}\textsuperscript{2}},
    \mbox{\textbf{Tianxing Zhou}\textsuperscript{8,5}},
    \mbox{\textbf{Zefang Huang}\textsuperscript{9,5}},
    \mbox{\textbf{Kang Chen}\textsuperscript{10,5}},
    \mbox{\textbf{Zhen Guo}\textsuperscript{11}},
    \mbox{\textbf{Quanlu Zhang}\textsuperscript{11}},
    \mbox{\textbf{Jincheng Yu}\textsuperscript{2}},
    \mbox{\textbf{Xiaodan Liang}\textsuperscript{6}},
    \mbox{\textbf{Guoliang Fan}\textsuperscript{1}},
    \mbox{\textbf{Yu Wang}\textsuperscript{2}},
    \mbox{\textbf{Feng Gao}\textsuperscript{2,3}},
    \mbox{\textbf{Xinlei Chen}\textsuperscript{2,\textdagger}},
    \mbox{\textbf{Chao Yu}\textsuperscript{2,\textdagger}}
    \\[8pt]
    {\normalfont\small
    \mbox{\textsuperscript{1}Institute of Automation, Chinese Academy of Sciences},
    \mbox{\textsuperscript{2}Tsinghua University},
    \mbox{\textsuperscript{3}Striding AI}, \\
    \mbox{\textsuperscript{4}School of Artificial Intelligence, University of Chinese Academy of Sciences}, \\
    \mbox{\textsuperscript{5}Zhongguancun Academy},
    \mbox{\textsuperscript{6}Pengcheng Laboratory},
    \mbox{\textsuperscript{7}Harbin Institute of Technology},
    \mbox{\textsuperscript{8}Beijing Institute of Technology},
    \mbox{\textsuperscript{9}Zhejiang University},
    \mbox{\textsuperscript{10}Peking University},
    \mbox{\textsuperscript{11}Infinigence AI}
    \\[5pt]
    \textsuperscript{*}Equal Contribution, \textsuperscript{\textdagger}Corresponding author
    }
  \end{minipage}
}

\iclrfinalcopy
\begin{document}

\maketitle

\begin{figure*}[h]
  \centering
  \includegraphics[width=1\linewidth]{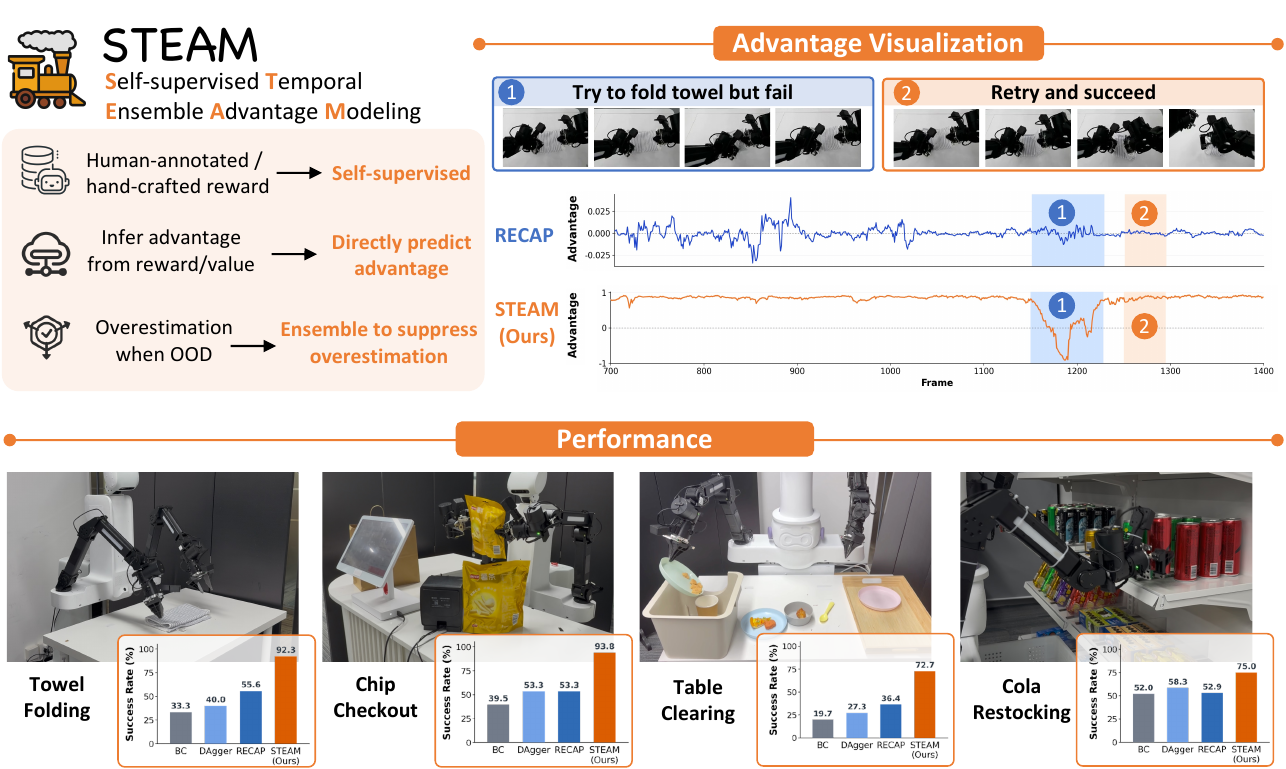}
    \caption{\textbf{Overview.} STEAM is a self-supervised advantage modeling framework for real-world robot learning. It learns advantage prediction offline from expert demonstrations without manual annotations or hand-crafted rewards, and can be applied to expert data, human corrections, and policy rollouts to provide robust advantage estimates. When combined with CFGRL~\citep{frans2025cfgrl}, STEAM substantially improves policy performance on various real-world tasks.}
  \label{fig:teaser}
\end{figure*}

\begin{abstract}
Real-world robot learning increasingly relies on heterogeneous data, but demonstrations and rollouts often mix useful progress with stalls, corrections, and suboptimal behavior.
Effective policy learning therefore requires frame-level advantages that distinguish reliable local progress from failures and regressions.
We propose \textbf{S}elf-supervised \textbf{T}emporal \textbf{E}nsemble \textbf{A}dvantage \textbf{M}odeling (\textbf{STEAM}), a label-free method that learns such advantages from expert demonstrations.
STEAM trains an ensemble of temporal-offset predictors on frame pairs within expert trajectories, using the normalized temporal offset between two frames as a self-supervised signal.
Each predictor maps a frame pair to a distribution over temporal offsets, which is converted into a scalar advantage.
STEAM then takes the minimum advantage across the ensemble to score mixed-quality rollout data conservatively.
Across real-world bimanual towel folding, chip checkout, table clearing, and cola restocking tasks, STEAM identifies stalls, failures, and recoveries.
When combined with CFGRL, STEAM achieves the highest policy success rate on all four tasks, averaging $83.5\%$ against $58.8\%$ for the best baseline.
Project page: \url{https://rlinf.github.io/steam/}.

\end{abstract}

\section{Introduction}
The rapid progress of robot foundation models relies heavily on large collections of expert demonstrations.
However, collecting expert trajectories is expensive, and real robot datasets often contain mixed-quality behavior, especially in complex long-horizon tasks~\citep{belkhale2023data, chen2026sarm, mao2026arm}.
To broaden state-action coverage and improve robustness, recent methods~\citep{amin2025recap, yang2026rise} incorporate non-expert trajectories, including policy rollouts, human interventions, and unsuccessful trials.
These trajectories can be useful, but their quality often changes within the same episode.
A rollout may make progress before failing, and a human-intervention trajectory may contain poor autonomous behavior before recovery.
Trajectory-level filtering can therefore discard useful transitions or keep harmful ones.
This raises a central question for policy improvement: \emph{how can we assign fine-grained credit to distinguish task-advancing actions from stalled or regressive ones?}

We formulate this question as frame-level advantage estimation.
Estimating such advantages without dense rewards remains difficult.
Recent robot-learning methods often follow a two-stage pipeline that first learns an intermediate reward or value signal~\citep{chen2026sarm, amin2025recap, yang2026rise,lee2026roboreward, liu2025timerewarder, zhai2025vlac}, then converts it into an advantage or sample weight for weighted behavior cloning~\citep{chen2026sarm,mao2026arm,xu2022discriminator} or offline reinforcement learning~\citep{frans2025cfgrl,kostrikov2022iql}.
However, deriving reliable signals remains difficult for three reasons.
First, hand-crafted rewards, human annotations, and cross-trajectory calibration require substantial external supervision and limit scalability.
Second, VLM-based reward or value models require broad pretraining and can still produce noisy frame-level signals when visual-language priors are not physically grounded.
Third, many progress estimators impose an absolute or monotonic notion of progress, which is poorly matched to real rollouts that may temporarily regress, recover after mistakes, or leave the expert distribution.
When a learned reward or value model is applied to these out-of-distribution rollout states, it may assign high advantages to unreliable transitions, creating false-positive training signals that drive policy optimization toward misleading behavior.

Instead of assigning an absolute progress value to each frame, we use the relative order of frames within the same expert trajectory.
This relative signal is local to each demonstration and avoids cross-trajectory calibration or a hand-specified global progress curve.
For any two frames, the normalized temporal offset can be computed directly from the trajectory.
This yields dense pairwise progress supervision without hand-crafted rewards, human labels, or external value estimates.

Building on such motivations, we introduce \textbf{STEAM} (\textbf{S}elf-supervised \textbf{T}emporal \textbf{E}nsemble \textbf{A}dvantage \textbf{M}odeling), a label-free framework for frame-level advantage estimation.
STEAM trains an ensemble of temporal-offset predictors on expert frame pairs.
Each predictor outputs a distribution over temporal bins, which is converted into a scalar advantage that reflects local temporal efficiency.
STEAM uses reversed frame pairs from successful trajectories to expose regressive motion without failed demonstrations.
To score mixed-quality trajectories robustly, STEAM takes the minimum advantage across the ensemble, suppressing overestimated advantages on out-of-distribution samples.
The resulting STEAM advantage can then be used to select high-advantage frames for policy improvement with offline RL methods such as classifier-free guidance reinforcement learning (CFGRL)~\citep{frans2025cfgrl}.
Our contributions are threefold.
\begin{itemize}[leftmargin=*, itemsep=0.4em]
\item
We formulate label-free frame-level advantage estimation from expert temporal structure, using normalized temporal offsets between frame pairs as dense self-supervised targets.
\item
We propose STEAM, which converts distributional temporal-offset predictions into scalar advantages and uses conservative ensemble aggregation to suppress overestimated advantages on mixed-quality data.
\item
We validate STEAM on real-world bimanual towel folding, chip checkout, table clearing, and cola restocking tasks.
STEAM localizes stalls, failures, and recoveries in demonstrations and rollouts, and when combined with CFGRL achieves the highest policy success rate on all four tasks, averaging $83.5\%$ against $58.8\%$ for the best baseline.
\end{itemize}

\section{Related Work}
\paragraph{Reward Models for Robot Manipulation.}
Refining policies from mixed-quality data requires fine-grained signals, such as rewards, from which value and advantage estimates can be derived to distinguish useful behaviors from suboptimal ones~\citep{peng2019awr,kostrikov2022iql}.
Since explicit rewards are rarely available in real-world robot manipulation, recent works instead learn reward or value
estimates from auxiliary supervision, including human annotations~\citep{chen2026sarm,mao2026arm}, hand-designed reward
or progress targets~\citep{amin2025recap,yang2026rise}, and foundation-model
priors~\citep{lee2026roboreward,tan2025robodopamine,ma2025gvl,liang2026robometer}.
In contrast, STEAM predicts advantage directly rather than recovering it from a
learned reward or value~\citep{amin2025recap,yang2026rise}.
This is similar to ARM~\citep{mao2026arm}, which estimates relative progress between
frame pairs. The key difference lies in supervision: ARM relies on a classifier trained on manually annotated data, whereas STEAM
exploits the temporal structure of expert demonstrations as self-supervision, requiring no manual annotation.

\paragraph{Temporal Structure as Supervision Signals.}
Temporal order is a free supervisory signal, since a frame's position along a trajectory encodes task progress.
This idea has long been used in self-supervised video representation learning~\citep{dwibedi2019tcc,zakka2022xirl,nair2022r3m,ma2023vip,ma2023liv}.
In robot learning, temporal order is typically converted into a dense reward signal to guide policy learning.
TimeRewarder~\citep{liu2025timerewarder} regresses frame-wise temporal offsets into dense rewards.
ReWiND~\citep{zhang2025rewind} rewinds successful videos into failure-like reward sequences.
VLAC~\citep{zhai2025vlac} uses a large vision-language critic for signed progress deltas.
Among existing methods, TimeRewarder is the closest prior work to STEAM, as both exploit signed temporal offsets between frame pairs, but STEAM uses them as per-frame advantages for offline data-quality assessment rather than dense rewards for online RL.

\begin{figure*}[t]
  \centering
  \includegraphics[width=1\linewidth]{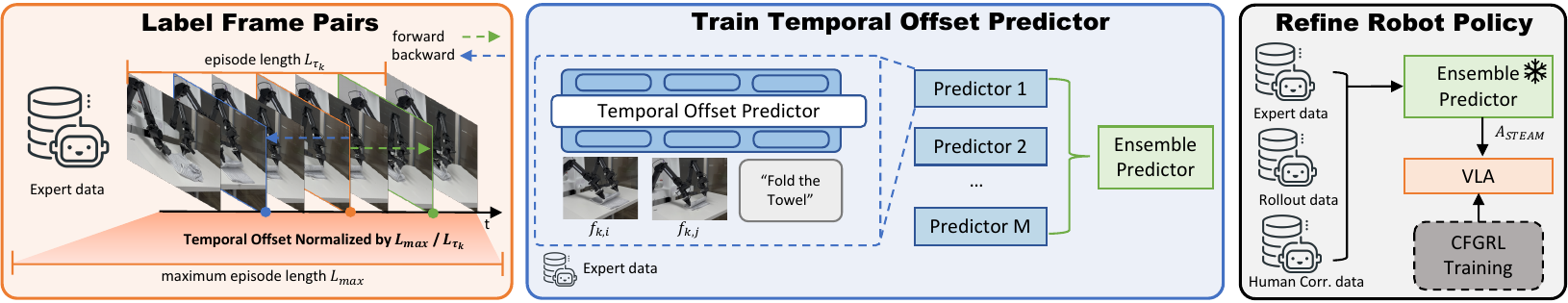}
  \caption{\textbf{STEAM framework.}
    (a) Expert demonstrations provide frame pairs for normalized temporal offset calculation. Both forward and reversed pairs are used as self-supervised targets.
    (b) An ensemble of $M$ predictors is trained on expert data to map frame pairs and language instructions to categorical distributions over temporal bins, converting these into scalar advantage scores.
    (c) The trained ensemble scores mixed-quality data, and these $\Asteam$ advantages then guide a VLA policy through CFGRL for robot policy refinement.
  }
  \label{fig:method}
\end{figure*}

\section{Methodology}
STEAM trains an ensemble of temporal-offset predictors using expert demonstrations.
These predictors then identify high-quality frames within heterogeneous, mixed-quality data to enhance policy performance (Figure~\ref{fig:method}).
Section~\ref{sec:pairwise} details our self-supervised temporal offset learning target.
Section~\ref{sec:value} turns those offsets into robust scalar advantages through a signed-bin predictor, and then into the binary optimality labels that policy training consumes.
Section~\ref{sec:theory} shows that the resulting score is an $H$-step advantage under an expert reference policy, so STEAM models the advantage directly instead of routing through a reward or a value model.

\subsection{Normalized Temporal Offset as Learning Target}
\label{sec:pairwise}

\paragraph{Self-Supervised Temporal Targets.}
Our training data comes exclusively from expert demonstrations.
Despite occasional brief stalls, the inherent temporal progression within these trajectories reliably indicates task advancement, eliminating the need for hand-crafted rewards.
For an expert episode $\tau_k = (f_{k,1}, f_{k,2}, \ldots, f_{k, L_{\tau_k}})$, we define the temporal offset between frames $f_{k,i}$ and $f_{k,j}$ as:
\begin{equation}
   \Delta_{\tau_k}(i, j) = \Delta(f_{k,i},\, f_{k,j}) = j-i.
  \label{eq:progress}
\end{equation}
Positive offsets, pairing a frame with a future observation, supervise forward progress.
Conversely, reversing expert trajectories provides negative offsets, creating pseudo-failure sequences that allow the model to learn regressive behaviors from successful demonstrations alone.

\paragraph{Trajectory-Length Normalization.}
Raw temporal offsets are not directly comparable across trajectories due to varying execution times.
A given offset $\Delta_{\tau_k}(i, j)$ might signify minor progress in a long episode but substantial progress in a short one.
To address this, we normalize the offset by trajectory length:
\begin{equation}
  \tilde\Delta_{\tau_k}(i, j)
  =
  \Delta(f_{k,i},\, f_{k,j}) \cdot \frac{L_{\max}}{L_{\tau_k}}
  =
  (j - i) \cdot \frac{L_{\max}}{L_{\tau_k}},
  \label{eq:ktilde}
\end{equation}
where $L_{\max}$ is the maximum length among expert episodes, or a high quantile of the episode lengths to reduce sensitivity to outliers.
This normalization is crucial as it unifies temporal-offset targets from diverse trajectories onto a consistent mathematical scale, ensuring stable supervision during training.

\subsection{From Offset Prediction to Optimality Labels}
\label{sec:value}

\paragraph{Categorical Prediction and Bin Assignment.}
We model the temporal-offset predictor as a distributional model. Given the long, high-frequency expert episodes, directly predicting continuous offsets would lead to an unmanageably large output space. To address this, we bound the continuous temporal offset $\tilde\Delta_{\tau_k}(i, j)$ within $[-\tilde\Delta_{\max}, \tilde\Delta_{\max}]$ and discretize it into $N$ evenly spaced bins. We write $\tilde\Delta^{\mathcal{B}}_{\tau_k}(i, j)$ for the index of the bin containing the offset, and use its one-hot encoding of length $N$ as the training target.
Our predictor, $p_\theta(\Delta \mid f_{k,i}, f_{k,j}, \ell)$, parameterized by $\theta$, maps an expert frame pair $(f_{k,i}, f_{k,j})$ and its natural language instruction $\ell$ to a categorical distribution over these $N$ temporal bins. We train this predictor by minimizing the cross-entropy loss:
\begin{equation}
  \mathcal{L}(\theta) \;=\;
  \mathbb{E}_{\tau_k,\,i,\,j}\!
  \left[
    \mathcal{H}\!\bigl(\,
    \tilde\Delta^{\mathcal{B}}_{\tau_k}(i, j),\;
      p_\theta(\Delta \mid f_{k,i}, f_{k,j}, \ell)
    \bigr)
  \right],
  \label{eq:loss}
\end{equation}
where $\mathcal{H}$ denotes the cross-entropy between the one-hot target and the predicted categorical distribution.

\paragraph{Advantage Modeling.}
To quantify a frame's contribution to task progress, we derive a scalar advantage from the learned distributional predictor.
Unlike standard RL advantages, our metric focuses solely on temporal efficiency.
For a frame $f_{k,i}$, we infer the model's prediction using a future frame $f_{k,i+H}$ (fixed lookahead $H$).
We then collapse the predicted categorical distribution into an expected bin index.
The advantage score is computed as the normalized expected bin index minus the ground-truth quantized temporal offset:
\begin{equation}
  A(f_{k,i};\, \theta)
  \;=\; \frac{2}{N}
  \left[
  \mathbb{E}_{b \sim p_\theta(\cdot \mid f_{k,i}, f_{k,i+H}, \ell)}[b]
  - \tilde\Delta^{\mathcal{B}}_{\tau_{\max}}(i, i+H)
  \right].
  \label{eq:advantage}
\end{equation}
Since $H$ is constant and $\tau_{\max}$ is the longest expert episode, $\tilde\Delta^{\mathcal{B}}_{\tau_{\max}}(i, i+H)$ is a fixed bin index. This formulation inherently prioritizes execution efficiency due to the trajectory-length normalization (Eq.~\plaineqref{eq:ktilde}). Shorter, more efficient expert demonstrations, scaled by a factor greater than one, yield higher advantage scores, while slower, suboptimal executions are penalized.

\paragraph{Ensemble for Overestimation Control.}
A single predictor trained on expert data can become overly confident on out-of-distribution rollout samples, leading to severely overestimated advantage scores~\citep{lakshminarayanan2017ensembles}.
Such overestimations generate false-positive signals that degrade policy refinement.
To mitigate this, we employ a conservative ensemble strategy.
We train $M$ independent temporal offset predictors $\Theta=\{\theta_1,\ldots,\theta_M\}$ with distinct random initializations using Equation~\plaineqref{eq:loss}.
Our final metric, the STEAM advantage, is the minimum predicted advantage across the ensemble:
\begin{equation}
  \Asteam(f_{k,i};\,\Theta)
  \;=\; \min_{m=1, \ldots, M}\; A(f_{k,i};\, \theta_m).
  \label{eq:ensemble}
\end{equation}
This minimum-aggregation exploits the tendency of ensemble members to agree within the training distribution but diverge in unfamiliar state spaces~\citep{lakshminarayanan2017ensembles}.
By penalizing high variance, this acts as a robust regularizer against reward overoptimization, a common issue in RLHF~\citep{coste2024ensembles}.
While this conservative bias might slightly reduce recall, it crucially suppresses false positives, which our empirical validation (Section~\ref{sec:ablation}) shows is vital for stable and effective policy optimization.

\paragraph{From Advantages to Optimality Labels.}
Policy training consumes $\Asteam$ through a binary optimality label. Since expert demonstrations and non-expert data (human corrections and autonomous rollouts) occupy distinct advantage ranges, we threshold each of the two pools separately rather than at a shared absolute value:
\begin{equation}
o_{k,i} = \mathbf{1}\left[\Asteam(f_{k,i};\Theta) \ge \delta_q\right],
\label{eq:label}
\end{equation}
where $\delta_q$ labels the top-$q$ fraction of each pool as optimal.
We supply $o_{k,i}$ as the optimality condition of CFGRL~\citep{frans2025cfgrl}, which steers action generation toward frames with high estimated local progress. Appendix~\ref{app:cfgrl} gives the full training and inference procedure.

\subsection{Theoretical Justification: Direct Advantage Modeling}
\label{sec:theory}

Although STEAM fits neither a reward model nor a value function, its score is an $H$-step advantage when elapsed time is the only cost, as we show next.

\paragraph{Reference value.}
Let every step before task completion receive the reward $r(s) = -1$, and let $\pi_{\mathrm{ref}}$ be a reference expert that completes the task in $L_{\max}$ steps.
Without discounting ($\gamma = 1$), the value $V_{\mathrm{ref}}(s)$ is the negative expected number of steps $\pi_{\mathrm{ref}}$ still needs from state $s$.

\paragraph{$H$-step advantage.}
For the $H$-step segment that starts at frame $f_{k,i}$, whose underlying state we denote by $s_{k,i}$, the undiscounted $H$-step advantage~\citep{schulman2016gae, amin2025recap} relative to $\pi_{\mathrm{ref}}$ is
\begin{equation}
  A^{(H)}(s_{k,i})
  \;=\; \underbrace{\sum\nolimits_{h=0}^{H-1} r(s_{k,i+h})}_{=\,-H}
  \;+\; \underbrace{V_{\mathrm{ref}}(s_{k,i+H}) - V_{\mathrm{ref}}(s_{k,i})}_{=\,\Delta V_{\mathrm{ref}}}
  \;=\; \Delta V_{\mathrm{ref}} - H.
  \label{eq:hstep}
\end{equation}
Since the reward term is the constant $-H$, estimating the advantage reduces to estimating the value difference $\Delta V_{\mathrm{ref}}$, which is the expected number of steps that $\pi_{\mathrm{ref}}$ needs to move from $s_{k,i}$ to $s_{k,i+H}$.

\paragraph{Temporal offsets as value-difference estimates.}
STEAM estimates this value difference directly instead of fitting $V_{\mathrm{ref}}$: an expert that moves from $s_{k,i}$ to $s_{k,j}$ takes $j-i$ of its own steps, and the rescaling by $L_{\max}/L_{\tau_k}$ in Equation~\plaineqref{eq:ktilde} converts them into steps of $\pi_{\mathrm{ref}}$, so that the normalized offset is a sampled estimate of the value difference:
\begin{equation}
  \tilde\Delta_{\tau_k}(i,j) \;=\; (j-i)\,\frac{L_{\max}}{L_{\tau_k}} \;\approx\; V_{\mathrm{ref}}(s_{k,j}) - V_{\mathrm{ref}}(s_{k,i}).
  \label{eq:offsetvalue}
\end{equation}
Deviations of individual episodes from the average pace average out over demonstrations, so the predictor trained with Equation~\plaineqref{eq:loss} estimates the value difference itself, much as sampled returns serve as targets for a value function in RECAP~\citep{amin2025recap}, but without fitting and then differencing a value function.

\paragraph{Correspondence to the STEAM advantage.}
In Equation~\plaineqref{eq:advantage}, $\mathbb{E}[b]$ estimates the bin of $\Delta V_{\mathrm{ref}}$ and the subtracted term is the bin of $H$, since the longest episode satisfies $\tilde\Delta_{\tau_{\max}}(i,i+H) = H$; as the bins have equal width $2\tilde\Delta_{\max}/N$,
\begin{equation*}
  A(f_{k,i};\theta)
  \;=\; \frac{2}{N}\Big[\mathbb{E}[b] - \tilde\Delta^{\mathcal{B}}_{\tau_{\max}}(i,i+H)\Big]
  \;\approx\; \frac{2}{N}\cdot\frac{N}{2\tilde\Delta_{\max}}\big(\Delta V_{\mathrm{ref}} - H\big)
  \;=\; \frac{A^{(H)}(s_{k,i})}{\tilde\Delta_{\max}},
\end{equation*}
that is, the STEAM advantage equals the $H$-step advantage scaled by the positive constant $1/\tilde\Delta_{\max}$.
In practice, we omit the constant term $\tilde\Delta^{\mathcal{B}}_{\tau_{\max}}(i,i+H)$, which does not change the labels in Equation~\plaineqref{eq:label}.

\section{Experiments}
\label{sec:experiments}

\paragraph{Tasks.}
We evaluate on four long-horizon bimanual tasks on an ARX dual-arm robot: towel folding, chip checkout, table clearing, and cola restocking.
They involve deformable objects, arm-to-arm handovers, and placement into tight spaces.
Appendices~\ref{app:hardware} and~\ref{app:task_details} give the robot setup, stage-by-stage procedure, and dataset composition of every task.

\paragraph{Baselines.}
We compare STEAM against four reward model baselines.
\textbf{ReWiND}~\citep{zhang2025rewind} rewinds successful demonstrations to synthesize failures, so its reward drops when the robot regresses.
\textbf{TimeRewarder}~\citep{liu2025timerewarder} learns a dense reward by regressing the temporal offset between video frames.
\textbf{RoboMeter}~\citep{liang2026robometer} uses a VLM as a general-purpose robotic reward model; we use the released RoboMeter-4B checkpoint.
\textbf{RECAP}~\citep{amin2025recap} uses task success and elapsed time as its reward.
The reward models of ReWiND and TimeRewarder are trained on expert demonstrations only. Each reward model then labels all expert and rollout frames, and these labels are used to train a value model and compute the advantage as in RECAP. STEAM is trained on expert demonstrations only and predicts the advantage directly, without fitting a value model. For policy learning, all methods label the same data and train the policy with CFGRL under identical settings.

\paragraph{Implementation.}
Each STEAM advantage predictor uses SigLIP-SO400M as the visual encoder and Gemma-3-270M as the language backbone, followed by task-specific prediction heads.
Hyperparameters are given in Appendix~\ref{app:hyperparams}.

We answer three questions:
\begin{itemize}[leftmargin=2.2em, itemsep=1pt, topsep=2pt, parsep=0pt]
  \item \textbf{Q1.} Does STEAM produce more robust advantage estimates than existing reward and value models on long-horizon, real-world manipulation data?
  \item \textbf{Q2.} Does STEAM improve real-world policy performance?
  \item \textbf{Q3.} How do STEAM's design choices and data composition affect final performance?
\end{itemize}

\subsection{Q1: Advantage Estimation Quality}
\label{sec:adv_eval}

\paragraph{Protocol and metrics.}
Evaluation consists of two parts.
First, \textbf{Clustering accuracy} tests whether the advantage statistically distinguishes between different successful and failed episodes.
Intuitively, successful episodes should accumulate higher advantage than failed ones.
We select 30 successful and 30 failed episodes and fit a two-component Gaussian mixture to divide them into two clusters by the sum of advantages.
Episodes in the higher-mean cluster are labeled $\hat{y}(\tau)=1$, otherwise $\hat{y}(\tau)=0$.
The clustering accuracy is the fraction of episodes classified correctly,
\begin{equation}
  \text{Clustering accuracy} = \frac{1}{|\mathcal{T}|}\sum_{\tau\in\mathcal{T}} \mathbf{1}\big[\hat{y}(\tau)= y(\tau)\big],
\end{equation}
where $y(\tau)$ is the ground-truth success label and $\mathcal{T}$ is the set of 60 episodes.
Second, \textbf{Intervention AUROC (I-AUROC)} tests whether a model detects an impending failure.
We use 20 human-correction episodes, in which an operator takes over when the policy is about to fail.
Around each takeover frame $t_{\mathrm{int}}$, we compare pre-takeover frames $\mathcal{F}^{-}$ with post-takeover frames $\mathcal{F}^{+}$,
\begin{equation}
  \mathcal{F}^{-}=[\,t_{\mathrm{int}}-H-W,\; t_{\mathrm{int}}-H\,),\qquad
  \mathcal{F}^{+}=[\,t_{\mathrm{int}}+5W,\; t_{\mathrm{int}}+6W\,),
\end{equation}
where $W=40$ (2\,s at 20\,Hz).
I-AUROC is the probability that a post-takeover frame receives a higher advantage than a pre-takeover frame,
\begin{equation}
  \text{I-AUROC} = \Pr\big[A(f_i) > A(f_j)\big], \quad f_i\in\mathcal{F}^{+},\; f_j\in\mathcal{F}^{-}.
\end{equation}

\paragraph{Results.}
Table~\ref{tab:adv_quality} compares the performance of the five models.
STEAM achieves the best results on both metrics, with 93.3\% clustering accuracy and 78.8\% I-AUROC, exceeding the strongest baseline by 5.0 and 13.7 points.
Its advantage drops as soon as the policy begins to fail and recovers once the task advances again, whereas the baselines respond weakly to failure; ReWiND, for example, stays near chance at 53.3\% I-AUROC, almost unable to distinguish pre- and post-takeover frames.

\begin{table}[!h]
\centering
\caption{\textbf{Advantage estimation quality.} Clust.\ Acc.: clustering accuracy (\%) of an unsupervised two-component Gaussian mixture on episode advantage sums. I-AUROC (\%): probability that a post-takeover frame receives a higher advantage than a pre-takeover frame.}
\label{tab:adv_quality}
\small
\begin{tabular}{lcc}
\toprule
\textbf{Method} & \textbf{Clust.\ Acc.\ $\uparrow$} & \textbf{I-AUROC $\uparrow$} \\
\midrule
ReWiND~\citep{zhang2025rewind} & 63.3 & 53.3 \\
TimeRewarder~\citep{liu2025timerewarder} & 88.3 & 65.1 \\
RoboMeter~\citep{liang2026robometer} & 75.0 & 45.5 \\
RECAP~\citep{amin2025recap} & 60.0 & 48.3 \\
\steamrow
STEAM (Ours) & \textbf{93.3} & \textbf{78.8} \\
\bottomrule
\end{tabular}
\end{table}

\begin{figure*}[t]
  \centering
  \includegraphics[width=1\linewidth]{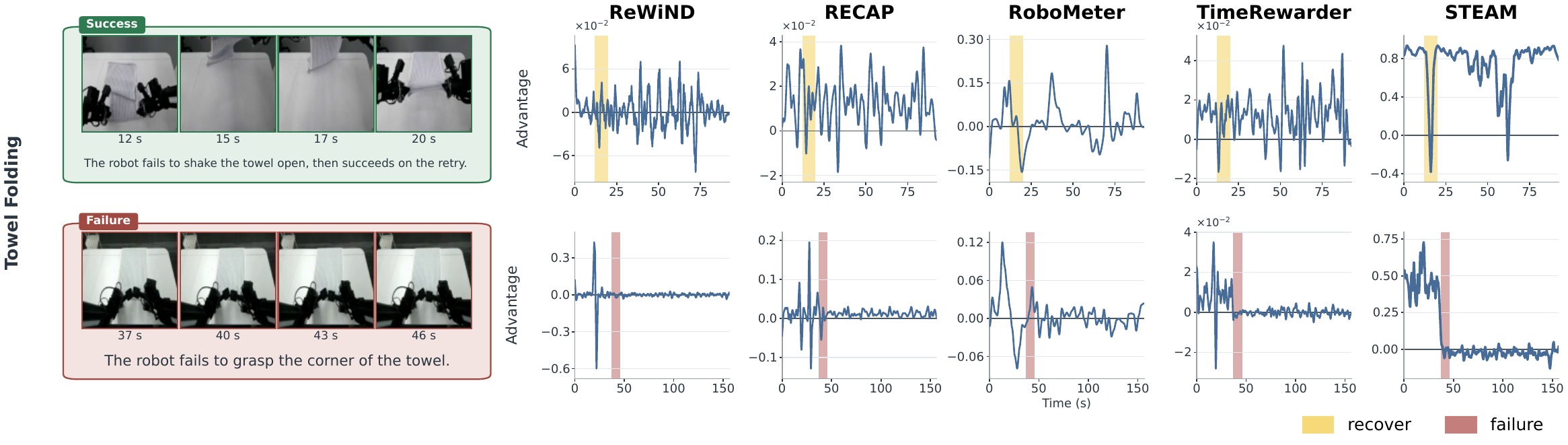}
  \caption{\textbf{Frame-level advantages on a successful and a failed towel-folding episode.}
  Each column shows the advantage assigned by one model to the same two episodes.
  In the successful episode (top), the robot fails to shake the towel open and then succeeds on the retry; the shaded band marks the recovery.
  In the failed episode (bottom), the robot never grasps the corner of the towel; the shaded band marks the onset of failure.
  Note the differing vertical scales.}
  \label{fig:adv_compare}
\end{figure*}

We further visualize the advantages of all models on a successful and a failed towel-folding episode in Figure~\ref{fig:adv_compare}.
STEAM stays at a stable positive level during normal execution.
In the successful episode, it dips at the retry and recovers; in the failed episode, it falls to zero once the robot stops making progress.
In contrast, the baselines fluctuate even during normal execution.
We attribute this difference to STEAM modeling the advantage directly, which reduces estimation error.
Appendix~\ref{app:viz} and~\ref{app:adv_dist} show more episodes and the corresponding advantage distributions.

\subsection{Q2: STEAM Enhances Policy Performance}
\label{sec:policy_eval}
We next ask whether better advantage estimates translate into better policies.
All methods label the same mixture of expert demonstrations, human corrections, and autonomous rollouts and train a $\pi_0$~\citep{black2024pi_0} policy through CFGRL for one iteration.

\begin{table*}[ht]
\centering
\caption{\textbf{Policy performance comparison.} Succ. represents the average success rate (\%), Score denotes the average number of completed sub-stages, Thr. is the throughput measured in successful episodes per hour, and Average is the mean success rate over the four tasks.
}
\label{tab:policy_perf}
\resizebox{\linewidth}{!}{
\begin{tabular}{lccccccccccccc}
\toprule
\textbf{Method} & \multicolumn{3}{c}{\textbf{Towel Folding}} & \multicolumn{3}{c}{\textbf{Chip Checkout}} & \multicolumn{3}{c}{\textbf{Table Clearing}} & \multicolumn{3}{c}{\textbf{Cola Restocking}} & \textbf{Average} \\
\cmidrule(lr){2-4} \cmidrule(lr){5-7} \cmidrule(lr){8-10} \cmidrule(lr){11-13} \cmidrule(lr){14-14}
& \textbf{Succ. (\%)} & \textbf{Score} & \textbf{Thr.} & \textbf{Succ. (\%)} & \textbf{Score} & \textbf{Thr.} & \textbf{Succ. (\%)} & \textbf{Score} & \textbf{Thr.} & \textbf{Succ. (\%)} & \textbf{Score} & \textbf{Thr.} & \textbf{Succ. (\%)} \\
\midrule
TimeRewarder & 53.3 & 3.9 & 27 & 70.6 & 5.7 & 30 & 46.7 & 14.5 & 20 & 64.7 & 2.6 & 64 & 58.8 \\
RoboMeter & 60.0 & 4.1 & 30 & 58.8 & 4.8 & 24 & 33.3 & 12.3 & 13 & 60.0 & 2.7 & 57 & 53.0 \\
RECAP & 55.6 & 2.9 & 39 & 53.3 & 5.3 & 24 & 36.4 & 12.5 & 15 & 52.9 & 2.1 & 46 & 49.6 \\
\steamrow
STEAM & \textbf{92.3} & \textbf{4.9} & \textbf{58} & \textbf{93.8} & \textbf{7.5} & \textbf{48} & \textbf{72.7} & \textbf{17.7} & \textbf{37.5} & \textbf{75.0} & \textbf{3.0} & \textbf{90} & \textbf{83.5} \\
\bottomrule
\end{tabular}
}
\end{table*}

As shown in Table~\ref{tab:policy_perf}, STEAM achieves the highest success rate on every task and raises the average success rate to 83.5\%, while the best baseline reaches only 58.8\%, and its margin is largest on the more complex tasks: STEAM leads the best baseline by 32.3 points on towel folding and 26.0 points on table clearing, which involves a deformable object and multi-step processes respectively. In contrast, on the simpler cola restocking task the margin narrows to 10.3 points.
We attribute this gap to the more stable advantage estimation of STEAM: the baselines must train a value model to obtain their advantages, whereas STEAM predicts the $H$-step advantage directly from two frames.
In complex tasks with frequent retries and stalls, such a value model is hard to fit, and its errors (Figure~\ref{fig:adv_compare}) lead the policy to imitate unproductive segments.
Conversely, on tasks composed of simple steps, the advantage of STEAM becomes less pronounced. Appendix~\ref{app:pnp_limitation} provides a further experiment on a short-horizon, two-stage task, where near-optimal expert data leave little improvement space until rollouts are added.

Moreover, STEAM does not trade speed for success: it also achieves the highest throughput, since it assigns lower advantages to segments executed more slowly than the expert (Equation~\plaineqref{eq:hstep}). We provide more visualizations in Appendix~\ref{app:viz}.

\subsection{Q3: Ablations on Design Choices and Data Composition}
\label{sec:ablation}
We first ablate the bin count $N$ and ensemble size $M$ on the towel-folding task with all expert demonstrations, human corrections, and rollout episodes, and then vary the composition of the training data.
Table~\ref{tab:ablation_bins} reports the effect of bin count, and Table~\ref{tab:ablation_ens} reports the effect of ensemble size.

\begin{table*}[ht]
\centering
\begin{minipage}[t]{0.48\textwidth}
\centering
\captionof{table}{Effect of bin count $N$.}
\label{tab:ablation_bins}
\small
\begin{tabular}{lccc}
\toprule
\textbf{Bins} & \textbf{Succ.} & \textbf{Score} & \textbf{Thr.} \\
\midrule
$N=2$  & 27.3 & 2.8 & 41 \\
$N=8$  & 54.6 & 3.8 & 51 \\
\rowcolor{gray!12}
$N=32$ (default) & \textbf{92.3} & \textbf{4.9} & \textbf{58} \\
\bottomrule
\end{tabular}
\end{minipage}
\hfill
\begin{minipage}[t]{0.48\textwidth}
\centering
\captionof{table}{Effect of ensemble size $M$.}
\label{tab:ablation_ens}
\small
\begin{tabular}{lccc}
\toprule
\textbf{Ens.} & \textbf{Succ.} & \textbf{Score} & \textbf{Thr.} \\
\midrule
$M=1$ & 72.7 & 3.9 & 53 \\
\rowcolor{gray!12}
$M=3$ (default) & \textbf{92.3} & \textbf{4.9} & \textbf{58} \\
$M=5$ & 90.9 & 4.6 & 55 \\
\bottomrule
\end{tabular}
\end{minipage}
\end{table*}

\paragraph{Effect of bin count.}
The bin count $N$ sets the resolution of the temporal-offset target.
With $N=2$, the target reduces to the sign of the offset, so every forward-moving segment receives nearly the same advantage regardless of its pace.
This coarse signal even falls below behavior cloning on towel folding (27.3\% against 33.3\% in Figure~\ref{fig:chart_bar}): the labels discard frames without ranking them by quality.
Finer bins let STEAM separate fast progress from slow progress, raising the success rate to 54.6\% at $N=8$ and 92.3\% at $N=32$, and throughput from 41 to 58 successful episodes per hour, consistent with the finer target distinguishing execution pace.

\begin{figure}[t]
  \centering
  \includegraphics[width=0.75\linewidth]{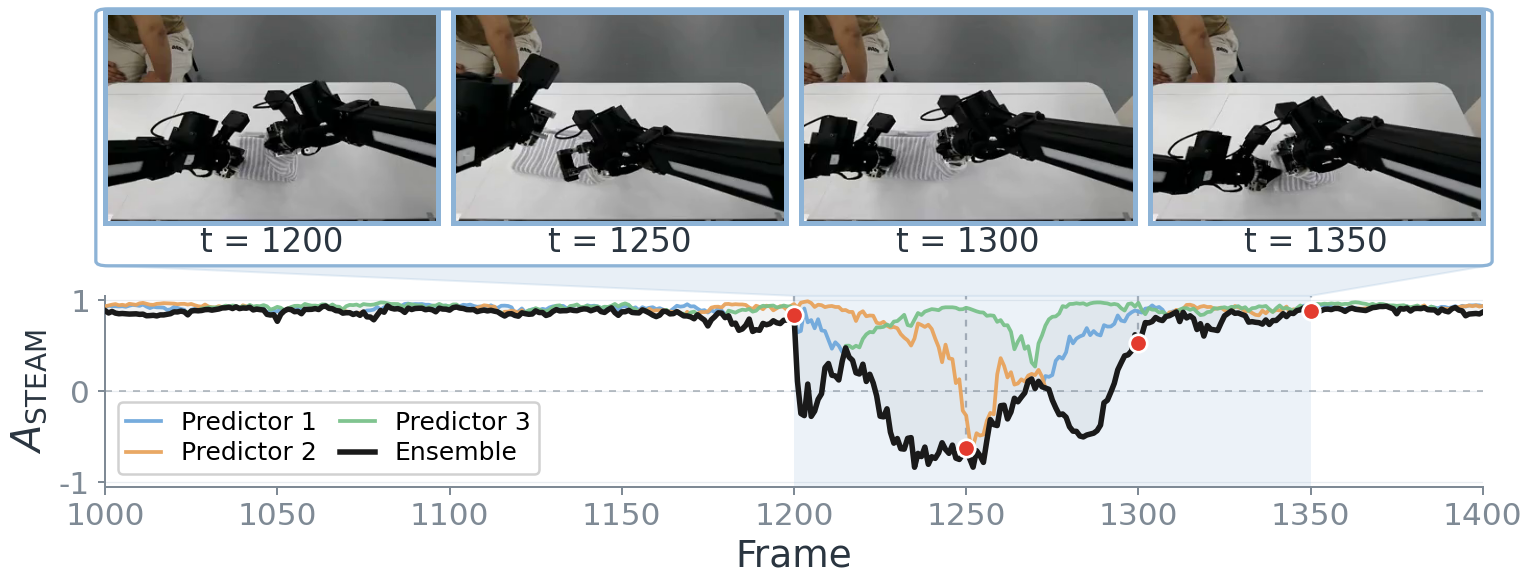}
  \caption{\textbf{Effect of conservative ensemble-min aggregation.} Advantages of the three ensemble members and the aggregated $A_{\mathrm{STEAM}}$ (black) on a towel-folding rollout. During the retry of the final fold (shaded, frames 1200--1350), Predictor 3 (green) stays near $+1$, while the minimum drops below $-0.5$.}
  \label{fig:ensemble_vis}
\end{figure}

\begin{figure*}[htb]
  \centering
  \includegraphics[width=1\linewidth]{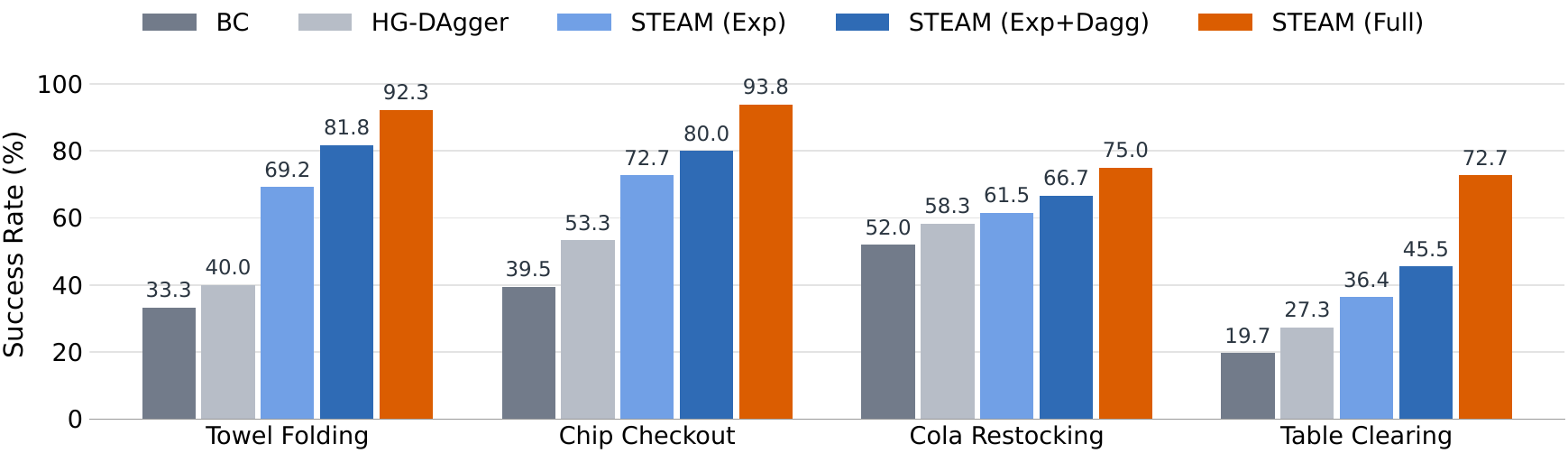}
  \caption{\textbf{STEAM performance across different training data combinations.} We compare the success rate of Behavior Cloning and HG-DAgger against STEAM trained with varying sources of data: expert demonstrations only (STEAM (Exp)), expert data supplemented with human correction episodes (STEAM (Exp+Dagg)), and the full dataset containing expert, correction, and autonomous rollout data (STEAM (Full)).}
  \label{fig:chart_bar}
\end{figure*}

\paragraph{Effect of ensemble size.}
The ensemble size $M$ controls how conservative the worst-of-$M$ aggregation is.
A single predictor ($M=1$) cannot flag its own errors on out-of-distribution rollout frames, so overestimated advantages admit poor segments into the optimal set and the success rate drops to 72.7\%.
With $M=3$, disagreement among the members pulls these frames down, and the success rate rises to 92.3\%.
A fifth member brings no further gain (90.9\%), so we use $M=3$ and avoid the extra training cost.
With a single predictor, such false positives appear as in Figure~\ref{fig:ensemble_vis}: Predictor 3 keeps a high advantage during a retry that Predictors 1 and 2 flag as regression, and the ensemble minimum suppresses it.

\paragraph{Effect of training data composition.}
To verify the effectiveness of STEAM on data from different sources, we further ablate the composition of the training data, from expert demonstrations only (STEAM (Exp)) to expert demonstrations with human corrections (STEAM (Exp+Dagg)) and the full mixture with autonomous rollouts (STEAM (Full)), and compare each setting against behavior cloning (BC) and HG-DAgger~\citep{kelly2019hg} in Figure~\ref{fig:chart_bar}.
STEAM effectively selects high-quality frames from both human corrections and autonomous rollouts, which cover states that expert demonstrations rarely visit, so adding these sources raises the success rate over STEAM (Exp) on all four tasks.
Even in the expert-only setting, STEAM (Exp) improves over both BC and HG-DAgger on all four tasks, although HG-DAgger additionally trains on human corrections, showing that STEAM isolates and emphasizes the task-advancing segments within expert demonstrations. The only exception is a short-horizon task with clean but limited expert data, analyzed in Appendix~\ref{app:pnp_limitation}, where filtering expert data alone slightly underperforms BC until rollouts and corrections are added.

\section{Limitations}
\label{sec:limitations}
STEAM learns advantages through temporal offsets, where frame pairs with the same offset receive the same target value. This ignores that different task phases may contribute unequally to progress. Structure-aware methods such as SPARS~\citep{he2026spars} assign different reward signals to different task stages, suggesting a possible way to incorporate phase-aware progress for fine-grained evaluation. Also, STEAM mainly relies on visual observations, which may fail to capture subtle but critical state differences.
Adding robot state as a learning target could further expand STEAM's ability to retrieve low-quality frames beyond visually observable errors.

\section{Conclusion}
We presented STEAM, a self-supervised framework that learns frame-level advantages from expert demonstrations alone, without hand-crafted rewards, annotations, or a fitted value model.
By training an ensemble on length-normalized temporal offsets, STEAM directly estimates an $H$-step advantage, which separates high- and low-quality segments across expert, corrected, and rollout data.
Its advantage separates successful from failed episodes with 93.3\% clustering accuracy and detects impending failures with 78.8\% I-AUROC, exceeding the strongest baseline by 5.0 and 13.7 points, respectively.
Combined with CFGRL on four real-world bimanual tasks, it achieves the highest success rate and throughput on every task, with the largest gains on the most complex ones.
These results suggest that estimating the advantage directly from temporal progress is a more reliable route to learning from mixed-quality robot data than fitting a reward model and then a value function.
Ablations show that fine-grained bins and the ensemble are both essential, and that every added data source, even of lower quality, further improves the policy.
Future work includes phase-aware progress and robot state to capture errors beyond visual observation.

\section*{AI Use Statement}

In this work, we used generative AI tools to clean and reformat datasets, to assist with
translation, to create and edit software code, and to draft and edit parts of this paper
for readability. We have not used generative AI tools to generate synthetic data sets, to
design or provide feedback on our research methodology or experiments, to implement the
proposed method, or to analyze or interpret our experimental results. The conceptual
framework of Section~\ref{sec:theory}, its mathematical claims, and the critical
ingredients of their derivation were developed by the authors.

We have reviewed all AI-assisted work. All LLM-generated code was read, tested, and
verified for correctness by the authors, and every experimental number reported in this
paper comes from real-robot runs of our own pipeline rather than from any AI-generated
content. All AI-drafted or AI-edited text was checked sentence by sentence by the
authors to ensure that the technical claims are accurate and reflect what was actually
done. All references were manually verified to exist and to be cited appropriately. We
take responsibility for the final content of this work, including text, claims or
artifacts produced with the aid of generative AI.

\clearpage

\bibliography{references}

@article{amin2025recap,
  title={$\pi^*_{0.6}$: a {VLA} That Learns From Experience},
  author={Amin, Ali and Aniceto, Raichelle and Balakrishna, Ashwin and
          Black, Kevin and Conley, Ken and Connors, Grace and Darpinian, James and
          Dhabalia, Karan and DiCarlo, Jared and Driess, Danny and others},
  journal={arXiv preprint arXiv:2511.14759},
  year={2025}
}

@article{mao2026arm,
  title={{ARM}: Advantage Reward Modeling for Long-Horizon Manipulation},
  author={Mao, Yiming and Yu, Zixi and Mao, Weixin and Li, Yinhao and
          Hu, Qirui and Lan, Zihan and Zhu, Minzhao and Chen, Hua},
  journal={arXiv preprint arXiv:2604.03037},
  year={2026}
}

@inproceedings{chen2026sarm,
  title={{SARM}: Stage-Aware Reward Modeling for Long Horizon Robot Manipulation},
  author={Chen, Qianzhong and Yu, Justin and Schwager, Mac and Abbeel, Pieter and
          Shentu, Yide and Wu, Philipp},
  booktitle={International Conference on Learning Representations},
  year={2026},
  url={https://openreview.net/forum?id=aemqAxScl9}
}

@inproceedings{zhang2025rewind,
  title={{ReWiND}: Language-Guided Rewards Teach Robot Policies without New Demonstrations},
  author={Zhang, Jiahui and Luo, Yusen and Anwar, Abrar and Sontakke, Sumedh A. and
          Lim, Joseph J. and Thomason, Jesse and B{\i}y{\i}k, Erdem and Zhang, Jesse},
  booktitle={9th Conference on Robot Learning},
  year={2025}
}

@inproceedings{ma2023vip,
  title={{VIP}: Towards Universal Visual Reward and Representation via
         Value-Implicit Pre-Training},
  author={Ma, Yecheng Jason and Sodhani, Shagun and Jayaraman, Dinesh and
          Bastani, Osbert and Kumar, Vikash and Zhang, Amy},
  booktitle={International Conference on Learning Representations},
  year={2023},
  url={https://openreview.net/forum?id=YJ7o2wetJ2}
}

@inproceedings{ma2023liv,
  title={{LIV}: Language-Image Representations and Rewards for Robotic Control},
  author={Ma, Yecheng Jason and Kumar, Vikash and Zhang, Amy and Bastani, Osbert and
          Jayaraman, Dinesh},
  booktitle={Proceedings of the 40th International Conference on Machine Learning},
  series={Proceedings of Machine Learning Research},
  volume={202},
  pages={23301--23320},
  year={2023}
}

@inproceedings{ma2025gvl,
  title={Vision Language Models are In-Context Value Learners},
  author={Ma, Yecheng Jason and Hejna, Joey and Wahid, Ayzaan and Fu, Chuyuan and
          Shah, Dhruv and Liang, Jacky and Xu, Zhuo and Kirmani, Sean and Xu, Peng and
          Driess, Danny and Xiao, Ted and Tompson, Jonathan and Bastani, Osbert and
          Jayaraman, Dinesh and Yu, Wenhao and Zhang, Tingnan and Sadigh, Dorsa and Xia, Fei},
  booktitle={International Conference on Learning Representations},
  year={2025},
  url={https://openreview.net/forum?id=friHAl5ofG}
}

@inproceedings{kostrikov2022iql,
  title={Offline Reinforcement Learning with Implicit Q-Learning},
  author={Kostrikov, Ilya and Nair, Ashvin and Levine, Sergey},
  booktitle={International Conference on Learning Representations},
  year={2022}
}

@inproceedings{kelly2019hg,
  title={{HG-DAgger}: Interactive imitation learning with human experts},
  author={Kelly, Michael and Sidrane, Chelsea and Driggs-Campbell, Katherine and Kochenderfer, Mykel J},
  booktitle={2019 International Conference on Robotics and Automation (ICRA)},
  pages={8077--8083},
  year={2019},
  organization={IEEE}
}

@inproceedings{lakshminarayanan2017ensembles,
  title={Simple and Scalable Predictive Uncertainty Estimation using Deep Ensembles},
  author={Lakshminarayanan, Balaji and Pritzel, Alexander and Blundell, Charles},
  booktitle={Advances in Neural Information Processing Systems},
  volume={30},
  year={2017}
}

@inproceedings{coste2024ensembles,
  title={Reward Model Ensembles Help Mitigate Overoptimization},
  author={Coste, Thomas and Anwar, Usman and Kirk, Robert and Krueger, David},
  booktitle={International Conference on Learning Representations},
  year={2024},
  url={https://arxiv.org/abs/2310.02743}
}

@article{frans2025cfgrl,
  title={Diffusion Guidance Is a Controllable Policy Improvement Operator},
  author={Frans, Kevin and Park, Seohong and Abbeel, Pieter and Levine, Sergey},
  journal={arXiv preprint arXiv:2505.23458},
  year={2025},
  url={https://arxiv.org/abs/2505.23458}
}

@inproceedings{dwibedi2019tcc,
  title={Temporal Cycle-Consistency Learning},
  author={Dwibedi, Debidatta and Aytar, Yusuf and Tompson, Jonathan and
          Sermanet, Pierre and Zisserman, Andrew},
  booktitle={Proceedings of the IEEE/CVF Conference on Computer Vision and
             Pattern Recognition},
  pages={1801--1810},
  year={2019},
  url={https://openaccess.thecvf.com/content_CVPR_2019/html/Dwibedi_Temporal_Cycle-Consistency_Learning_CVPR_2019_paper.html}
}

@inproceedings{zakka2022xirl,
  title={{XIRL}: Cross-Embodiment Inverse Reinforcement Learning},
  author={Zakka, Kevin and Zeng, Andy and Florence, Pete and Tompson, Jonathan and
          Bohg, Jeannette and Dwibedi, Debidatta},
  booktitle={5th Conference on Robot Learning},
  year={2022},
  url={https://openreview.net/forum?id=RO4DM85Z4P7}
}

@inproceedings{nair2022r3m,
  title={{R3M}: A Universal Visual Representation for Robot Manipulation},
  author={Nair, Suraj and Rajeswaran, Aravind and Kumar, Vikash and
          Finn, Chelsea and Gupta, Abhinav},
  booktitle={6th Conference on Robot Learning},
  year={2022},
  url={https://arxiv.org/abs/2203.12601}
}

@article{zhai2025vlac,
  title={A Vision-Language-Action-Critic Model for Robotic Real-World Reinforcement Learning},
  author={Zhai, Shaopeng and Zhang, Qi and Zhang, Tianyi and Huang, Fuxian and Zhang, Haoran and Zhou, Ming and Zhang, Shengzhe and Liu, Litao and Lin, Sixu and Pang, Jiangmiao},
  journal={arXiv preprint arXiv:2509.15937},
  year={2025},
  url={https://arxiv.org/abs/2509.15937}
}

@article{peng2019awr,
  title={Advantage-Weighted Regression: Simple and Scalable Off-Policy Reinforcement Learning},
  author={Peng, Xue Bin and Kumar, Aviral and Zhang, Grace and Levine, Sergey},
  journal={arXiv preprint arXiv:1910.00177},
  year={2019},
  url={https://arxiv.org/abs/1910.00177}
}

@article{lee2026roboreward,
  title={{RoboReward}: General-Purpose Vision-Language Reward Models for Robotics},
  author={Lee, Tony and Wagenmaker, Andrew and Pertsch, Karl and Liang, Percy and Levine, Sergey and Finn, Chelsea},
  journal={arXiv preprint arXiv:2601.00675},
  year={2026},
  url={https://arxiv.org/abs/2601.00675}
}

@inproceedings{liang2026robometer,
  title={Robometer: Scaling General-Purpose Robotic Reward Models via Trajectory Comparisons},
  author={Liang, Anthony and Korkmaz, Yigit and Zhang, Jiahui and Hwang, Minyoung and Anwar, Abrar and Kaushik, Sidhant and Shah, Aditya and Huang, Alex S. and Zettlemoyer, Luke and Fox, Dieter and Xiang, Yu and Li, Anqi and Bobu, Andreea and Gupta, Abhishek and Tu, Stephen and Biyik, Erdem and Zhang, Jesse},
  booktitle={Robotics: Science and Systems},
  year={2026},
  url={https://arxiv.org/abs/2603.02115}
}

@inproceedings{tan2025robodopamine,
  title={General Process Reward Modeling for Robotic Reinforcement Learning},
  author={Tan, Huajie and Chen, Sixiang and Xu, Yijie and Wang, Zixiao and Chi, Cheng and Ji, Yuheng and Lyu, Yaoxu and Zhao, Zhongxia and Chen, Xiansheng and Co, Peterson and Xie, Shaoxuan and Yao, Guocai and Wang, Pengwei and Wang, Zhongyuan and Zhang, Shanghang},
  booktitle={Proceedings of the IEEE/CVF Conference on Computer Vision and Pattern Recognition},
  pages={22412--22422},
  year={2026},
  url={https://arxiv.org/abs/2512.23703}
}

@inproceedings{yang2026rise,
  title={{RISE}: Self-Improving Robot Policy with Compositional World Model},
  author={Yang, Jiazhi and Lin, Kunyang and Li, Jinwei and Zhang, Wencong and Lin, Tianwei and Wu, Longyan and Su, Zhizhong and Zhao, Hao and Zhang, Ya-Qin and Chen, Li and Luo, Ping and Yue, Xiangyu and Li, Hongyang},
  booktitle={Robotics: Science and Systems},
  year={2026},
  url={https://arxiv.org/abs/2602.11075}
}

@article{belkhale2023data,
  title={Data quality in imitation learning},
  author={Belkhale, Suneel and Cui, Yuchen and Sadigh, Dorsa},
  journal={Advances in neural information processing systems},
  volume={36},
  pages={80375--80395},
  year={2023}
}

@inproceedings{black2024pi_0,
  title={$\pi_0$: A Vision-Language-Action Flow Model for General Robot Control},
  author={Black, Kevin and Brown, Noah and Driess, Danny and Esmail, Adnan and Equi, Michael and Finn, Chelsea and Fusai, Niccolo and Groom, Lachy and Hausman, Karol and Ichter, Brian and others},
  booktitle={Robotics: Science and Systems},
  year={2025}
}

@inproceedings{xu2022discriminator,
  title={Discriminator-weighted offline imitation learning from suboptimal demonstrations},
  author={Xu, Haoran and Zhan, Xianyuan and Yin, Honglei and Qin, Huiling},
  booktitle={International Conference on Machine Learning},
  pages={24725--24742},
  year={2022},
  organization={PMLR}
}

@inproceedings{liu2025timerewarder,
  title={TimeRewarder: Learning Dense Reward from Passive Videos via Frame-wise Temporal Distance},
  author={Liu, Yuyang and Wen, Chuan and Hu, Yihang and Jayaraman, Dinesh and Gao, Yang},
  booktitle={International Conference on Machine Learning},
  year={2026}
}

@article{he2026spars,
  title={SPARS: Structure-informed progress-aware reward shaping for fabric manipulation learning from demonstration},
  author={He, Ruhan and Wei, Yifei and Yu, Lianqing and Zeng, Xianyi},
  journal={Robotics and Autonomous Systems},
  volume={203},
  pages={105499},
  year={2026},
  publisher={Elsevier}
}

@inproceedings{schulman2016gae,
  title={High-Dimensional Continuous Control Using Generalized Advantage Estimation},
  author={Schulman, John and Moritz, Philipp and Levine, Sergey and Jordan, Michael and Abbeel, Pieter},
  booktitle={International Conference on Learning Representations},
  year={2016},
  url={https://arxiv.org/abs/1506.02438}
}
\bibliographystyle{iclr2027_conference}

\newpage
\appendix

\section{Appendix}
\label{app:impl}

\subsection{Hardware Setup}
\label{app:hardware}

\begin{figure}[ht]
  \centering
  \includegraphics[width=0.62\linewidth]{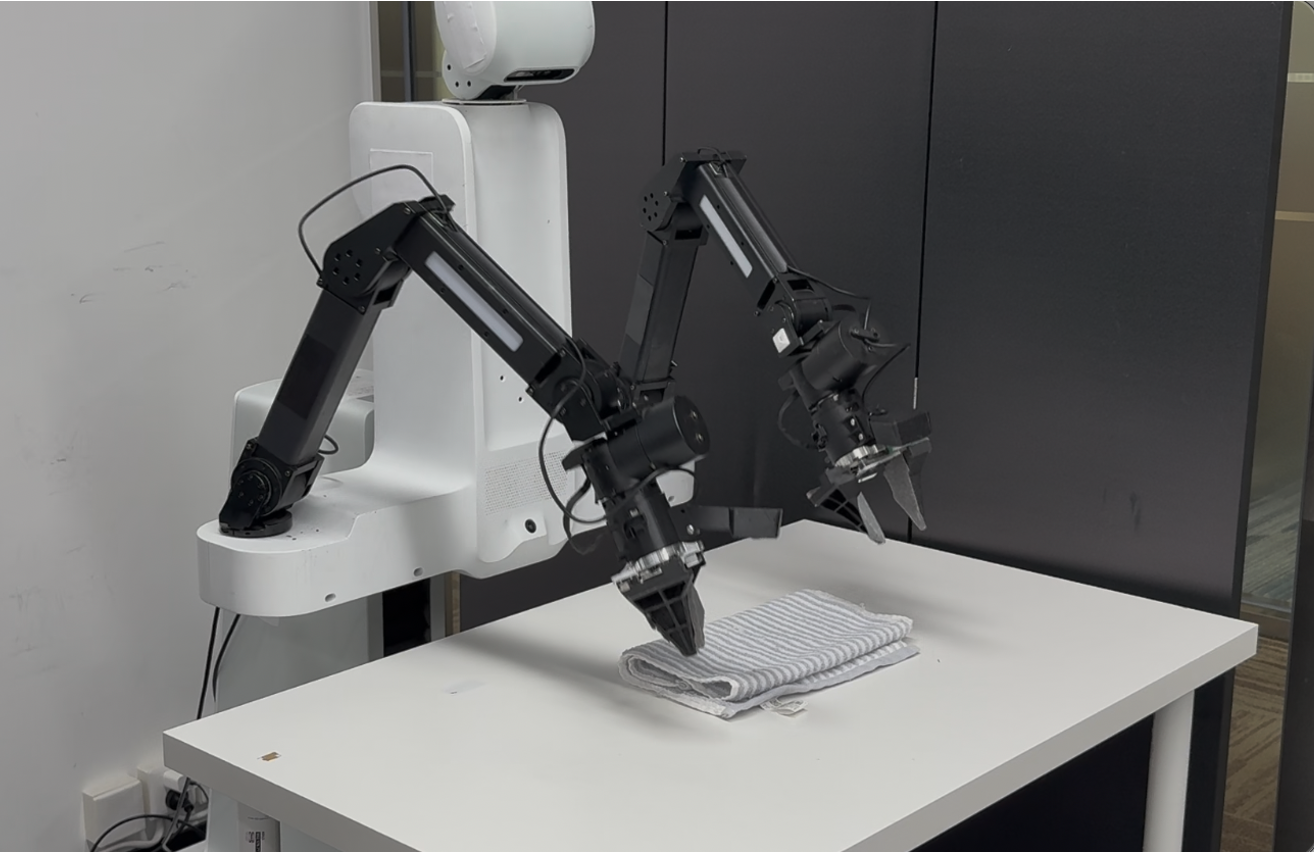}
  \caption{\textbf{The bimanual platform used for data collection and policy evaluation.}
  Two ARX arms are mounted on a fixed torso facing the table, each carrying a wrist camera, with a third
  camera on the head looking down on the workspace. The scene shown is the towel-folding task.}
  \label{fig:platform}
\end{figure}

All real-robot experiments are carried out on the bimanual tabletop platform shown in Figure~\ref{fig:platform}. The system consists of:
\begin{itemize}
    \item Two ARX robot arms, built by the manufacturer ARX Robotics, mounted side by side in a bimanual configuration.
    \item Three Orbbec Astra Pro RGB cameras: one mounted on each wrist and a third mounted on the head of the platform for an overview of the workspace.
\end{itemize}

Both expert demonstrations and human-correction episodes are collected through a leader--follower teleoperation system: the operator drives a pair of kinematically matched leader arms, and the follower arms on the platform reproduce the motion. The same interface is used during autonomous execution, which allows the operator to take over at any moment and hand control back once the robot has recovered; the frames recorded while the operator is in control are exactly the human-correction data used throughout this paper.

Episodes are recorded at $20$~fps and contain the three synchronized camera streams (head, left wrist, right wrist) together with the robot proprioception. We use the \emph{absolute} end-effector pose as both the state and the action representation, i.e.\ the Cartesian position and orientation of each gripper expressed in the robot base frame, rather than joint angles or relative displacements, so that the advantage model and the policy operate in the same task-space coordinates across all tasks.

The pick-and-place task analyzed in Appendix~\ref{app:pnp_limitation} is the one exception to this setup: it runs on a single Franka arm and is included as a short-horizon control condition.

\subsection{Task Descriptions and Training Data Composition}
\label{app:task_details}

\begin{figure}[htbp]
\centering
\includegraphics[width=1\linewidth]{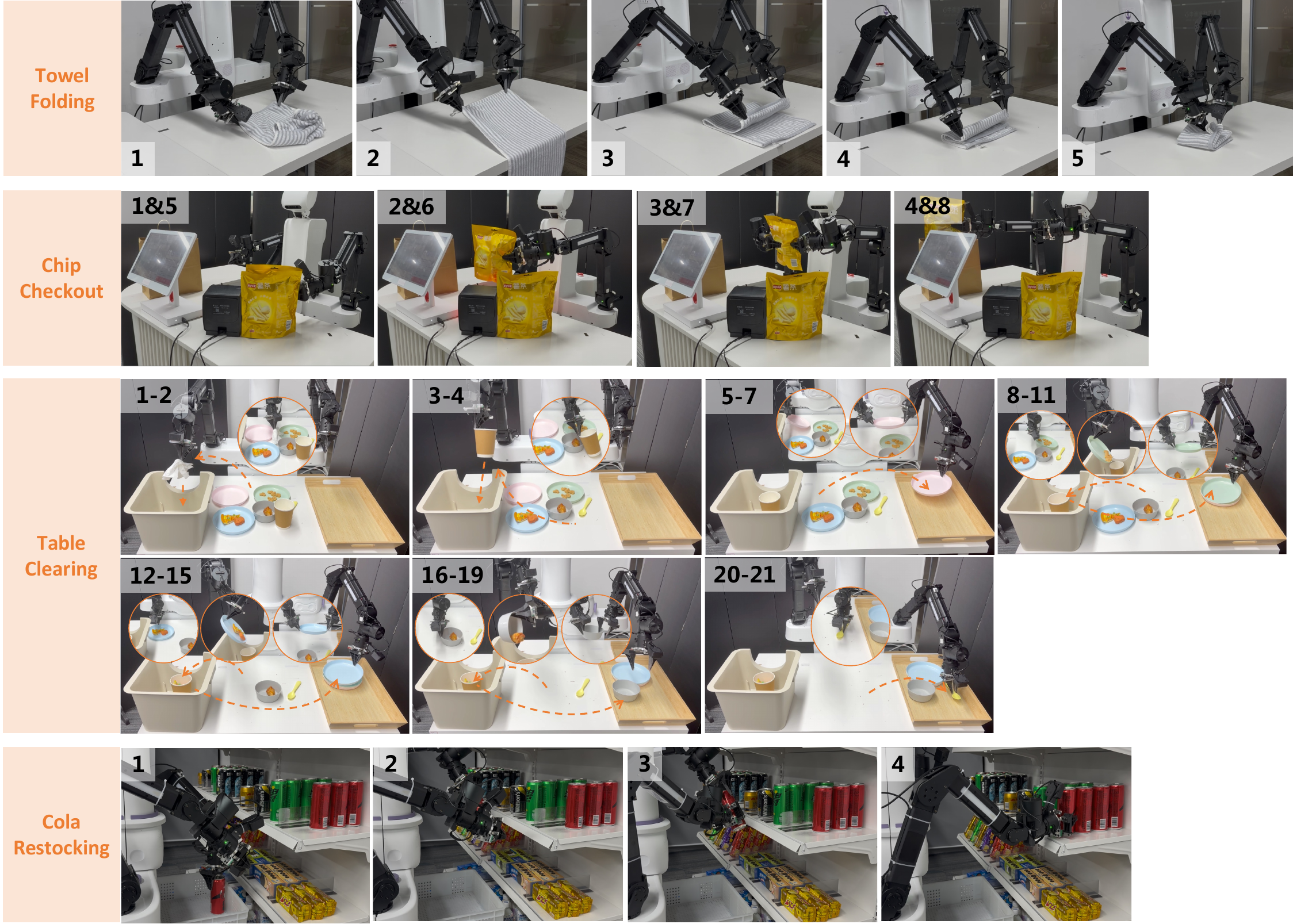}
\caption{\textbf{Robot setup and tasks.} The four bimanual tasks used throughout the paper --- towel folding, chip checkout, table clearing, and cola restocking --- with the stages of each task and the sub-step indices each panel covers.}
\label{fig:setup}
\end{figure}

To evaluate STEAM across different manipulation horizons, contact patterns, and coordination requirements, we design a set of real-world tasks, shown in Figure~\ref{fig:setup}. Towel folding (5 stages), chip checkout (8 stages), table clearing (21 stages), and cola restocking (4 stages) run on an ARX dual-arm robot. For each task we collect a dataset that mixes expert demonstrations, autonomous rollouts, and human correction episodes in varying proportions. Table~\ref{tab:data_composition} summarizes the resulting dataset composition, and the step-by-step procedure of each task is given below.

\begin{table}[ht]
  \caption{\textbf{Training data composition.} Number of episodes of each data source per task. Expert demonstrations are collected by teleoperation, autonomous rollouts are generated by a behavior-cloning policy trained on the expert data, and human-correction episodes are autonomous rollouts in which the operator took over and recovered the task.}
  \label{tab:data_composition}
  \centering
  \small
  \begin{tabular}{lccccc}
    \toprule
    \textbf{Task} & \textbf{Stages} & \textbf{Expert demos} & \textbf{Autonomous rollouts} & \textbf{Human corrections} & \textbf{Total} \\
    \midrule
    Towel Folding   & 5  & 240 & 125 & 20 & 385 \\
    Chip Checkout   & 8  & 200 & 63  & 20 & 283 \\
    Table Clearing  & 21 & 203 & 94  & 42 & 339 \\
    Cola Restocking & 4  & 89  & 49  & 27 & 165 \\
    \bottomrule
  \end{tabular}
\end{table}

\paragraph{Towel Folding.}
This long-horizon task utilizes an ARX dual-arm robot to fold a towel. A successful execution requires completing all five consecutive stages without severe fabric misfolding. The dataset collected for this task includes 240 expert demonstrations, 125 autonomous rollouts, and 20 human correction episodes. The folding sequence is divided into five stages:
\begin{itemize}
    \item \textbf{Stage 1 (Pick up):} The robot arms reach down and grasp the corners of the towel on the table.
    \item \textbf{Stage 2 (Flatten):} The arms lift and stretch the towel to flatten the fabric, preparing it for folding.
    \item \textbf{Stage 3 (First fold):} The robot executes the first longitudinal fold.
    \item \textbf{Stage 4 (Second fold):} The robot executes the second fold, reducing the towel's width.
    \item \textbf{Stage 5 (Third fold):} The robot performs the final fold to complete the folding sequence.
\end{itemize}

\paragraph{Chip Checkout.}
This dual-arm coordination task simulates a retail checkout process, where two bags of chips must be scanned and bagged sequentially. The training dataset for this task includes 200 expert demonstrations, 63 autonomous rollouts, and 20 human correction episodes. The task involves 8 stages:
\begin{itemize}
    \item \textbf{Stage 1 (Pick up first bag):} The left arm reaches and grasps the first bag of chips from the counter.
    \item \textbf{Stage 2 (Scan first bag):} The left arm moves the chip bag to the barcode scanner and performs the scanning action.
    \item \textbf{Stage 3 (Handover first bag):} The left arm coordinates with the right arm to handover the first chip bag.
    \item \textbf{Stage 4 (Bag first bag):} The right arm receives the chip bag and places it into the shopping bag.
    \item \textbf{Stage 5 (Pick up second bag):} The left arm reaches and grasps the second bag of chips.
    \item \textbf{Stage 6 (Scan second bag):} The left arm moves the second chip bag to the scanner for barcode recognition.
    \item \textbf{Stage 7 (Handover second bag):} The left arm handovers the second chip bag to the right arm.
    \item \textbf{Stage 8 (Bag second bag):} The right arm receives the second chip bag and completes the bagging process.
\end{itemize}

\paragraph{Table Clearing.}
This is the longest task in our suite: the dual-arm robot has to clear a dinner table of seven objects, discarding the disposable items, emptying the leftover food from every container, and stacking the containers on the side. It is also the most demanding one in terms of coordination, since every plate and bowl is handed over from the right arm to the left arm before being put down. The training dataset contains 203 expert demonstrations, 94 autonomous rollouts, and 42 human correction episodes. The task is divided into 21 stages:
\begin{itemize}
    \item \textbf{Stage 1 (Pick up tissue):} The right arm grasps the used tissue from the table.
    \item \textbf{Stage 2 (Discard tissue):} The right arm drops the tissue into the trash bin.
    \item \textbf{Stage 3 (Pick up paper cup):} The right arm grasps the paper cup.
    \item \textbf{Stage 4 (Discard paper cup):} The right arm drops the paper cup into the trash bin.
    \item \textbf{Stage 5 (Pick up pink plate):} The right arm grasps the pink plate.
    \item \textbf{Stage 6 (Handover pink plate):} The right arm hands the pink plate over to the left arm.
    \item \textbf{Stage 7 (Place pink plate):} The left arm places the pink plate onto the stacking area.
    \item \textbf{Stage 8 (Pick up green plate):} The right arm grasps the green plate.
    \item \textbf{Stage 9 (Empty green plate):} The right arm tips the leftover food from the green plate into the trash bin.
    \item \textbf{Stage 10 (Handover green plate):} The right arm hands the green plate over to the left arm.
    \item \textbf{Stage 11 (Place green plate):} The left arm places the green plate onto the stack.
    \item \textbf{Stage 12 (Reposition and pick up blue plate):} The right arm pushes the blue plate into a graspable pose and picks it up.
    \item \textbf{Stage 13 (Empty blue plate):} The right arm tips the leftover food from the blue plate into the trash bin.
    \item \textbf{Stage 14 (Handover blue plate):} The right arm hands the blue plate over to the left arm.
    \item \textbf{Stage 15 (Place blue plate):} The left arm places the blue plate onto the stack.
    \item \textbf{Stage 16 (Pick up grey bowl):} The right arm grasps the grey bowl.
    \item \textbf{Stage 17 (Empty grey bowl):} The right arm tips the leftover food from the grey bowl into the trash bin.
    \item \textbf{Stage 18 (Handover grey bowl):} The right arm hands the grey bowl over to the left arm.
    \item \textbf{Stage 19 (Place grey bowl):} The left arm places the grey bowl onto the stack.
    \item \textbf{Stage 20 (Pick up yellow spoon):} The right arm grasps the yellow spoon.
    \item \textbf{Stage 21 (Place yellow spoon):} The right arm places the yellow spoon at its target position, completing the task.
\end{itemize}

\paragraph{Cola Restocking.}
This task requires the dual-arm robot to transfer a cola bottle from a crate to a shelf, which demands active collision avoidance. The training dataset consists of 89 expert demonstrations, 27 human corrections, and 49 autonomous rollouts. The task is divided into 4 stages:
\begin{itemize}
    \item \textbf{Stage 1 (Pick up):} The left arm grasps a cola bottle from the inventory crate.
    \item \textbf{Stage 2 (Camera adjustment):} The left arm rotates its wrist to adjust the wrist-camera's viewpoint, preventing potential hardware and camera collisions between the two arms.
    \item \textbf{Stage 3 (Handover):} The left arm handovers the cola bottle to the right arm.
    \item \textbf{Stage 4 (Place):} The right arm moves the cola bottle and places it stably onto the target shelf.
\end{itemize}

\subsection{Per-Task Advantage Visualizations}
\label{app:viz}

Section~\ref{sec:adv_eval} compares STEAM against the baselines on two towel-folding episodes. Here we show the frame-level $A_{\mathrm{STEAM}}$ curve on all four tasks and on all four episode types we collect (expert demonstrations, successful rollouts, failed rollouts, and human-correction episodes) to document what the signal looks like across the full range of data STEAM is applied to.

An effective advantage signal must reflect the nuances of task execution: remaining high during proficient segments, decreasing during hesitation or failure, and recovering upon effective progress.

\begin{figure}[htbp]
  \centering
  \includegraphics[width=\linewidth]{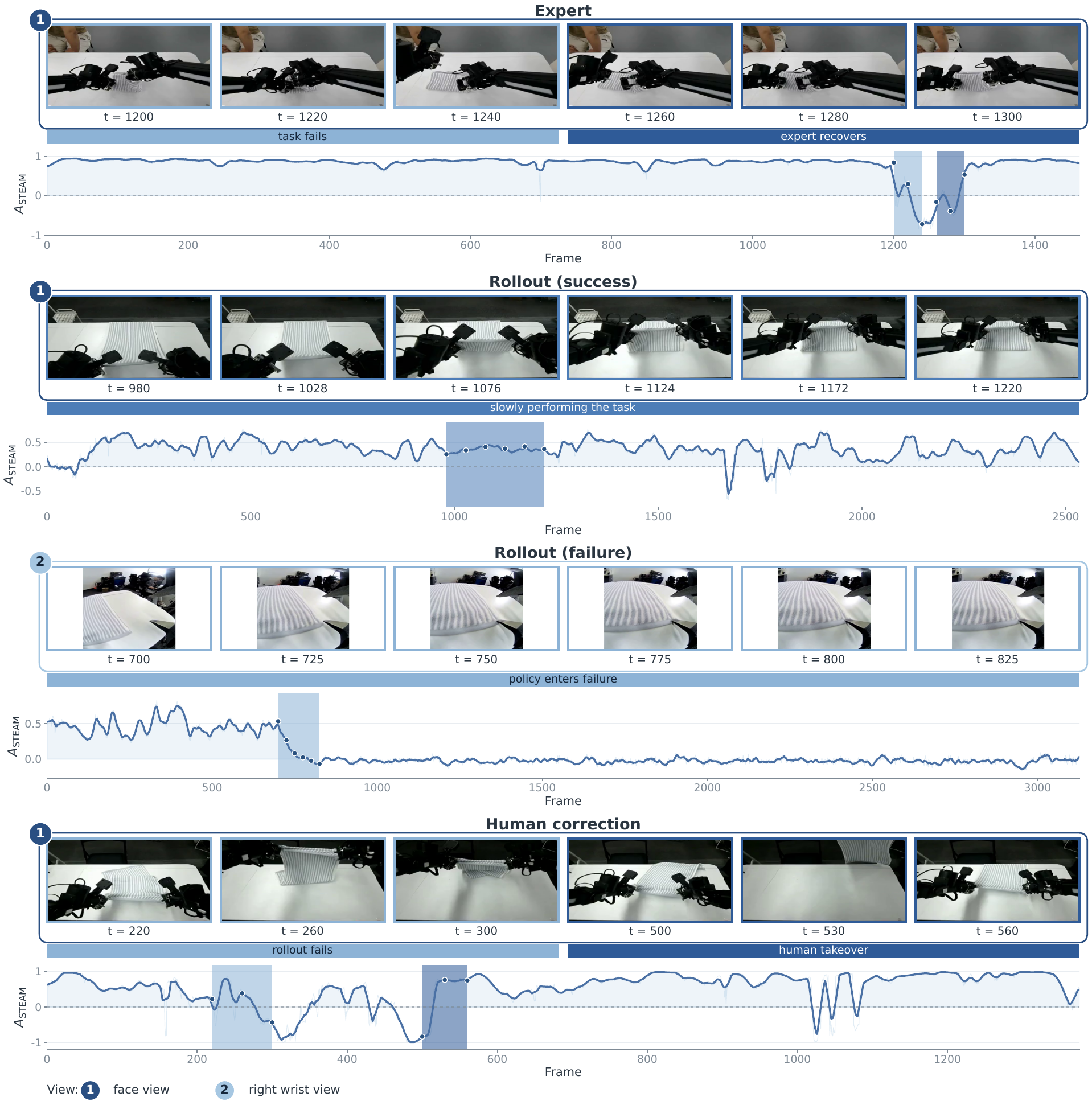}
  \caption{\textbf{Towel folding advantage curves.}
Frame-level $A_{\mathrm{STEAM}}$ on four representative episode types: expert demonstrations, successful rollouts, failed rollouts, and human correction episodes. Images on top show
corresponding frames from each episode, while shaded regions highlight
segments with retry, slow progress, failure, or human takeover.}
  \label{fig:fig4}
\end{figure}

\paragraph{Towel Folding.}
Figure~\ref{fig:fig4} visualizes the four representative episode types in the towel folding task.
For expert demonstrations, the advantage values remain consistently high throughout the majority of the episode, experiencing only minor and transient drops around frames that involve necessary retries or slight adjustments.
Successful rollouts show lower and more fluctuating advantages than expert demonstrations, reflecting slower yet still effective task execution.
In contrast, failed rollouts rapidly drop to near-zero advantage after entering failure states, where the policy becomes stuck without recovering.
Human correction episodes present an initial drop similar to failures, but critically, show a clear recovery of advantage after human intervention.
STEAM therefore provides a fine-grained, frame-level assessment capable of distinguishing proficient execution from hesitation, failure, and recovery within a single episode.

\begin{figure}[htbp]
  \centering
  \includegraphics[width=\linewidth]{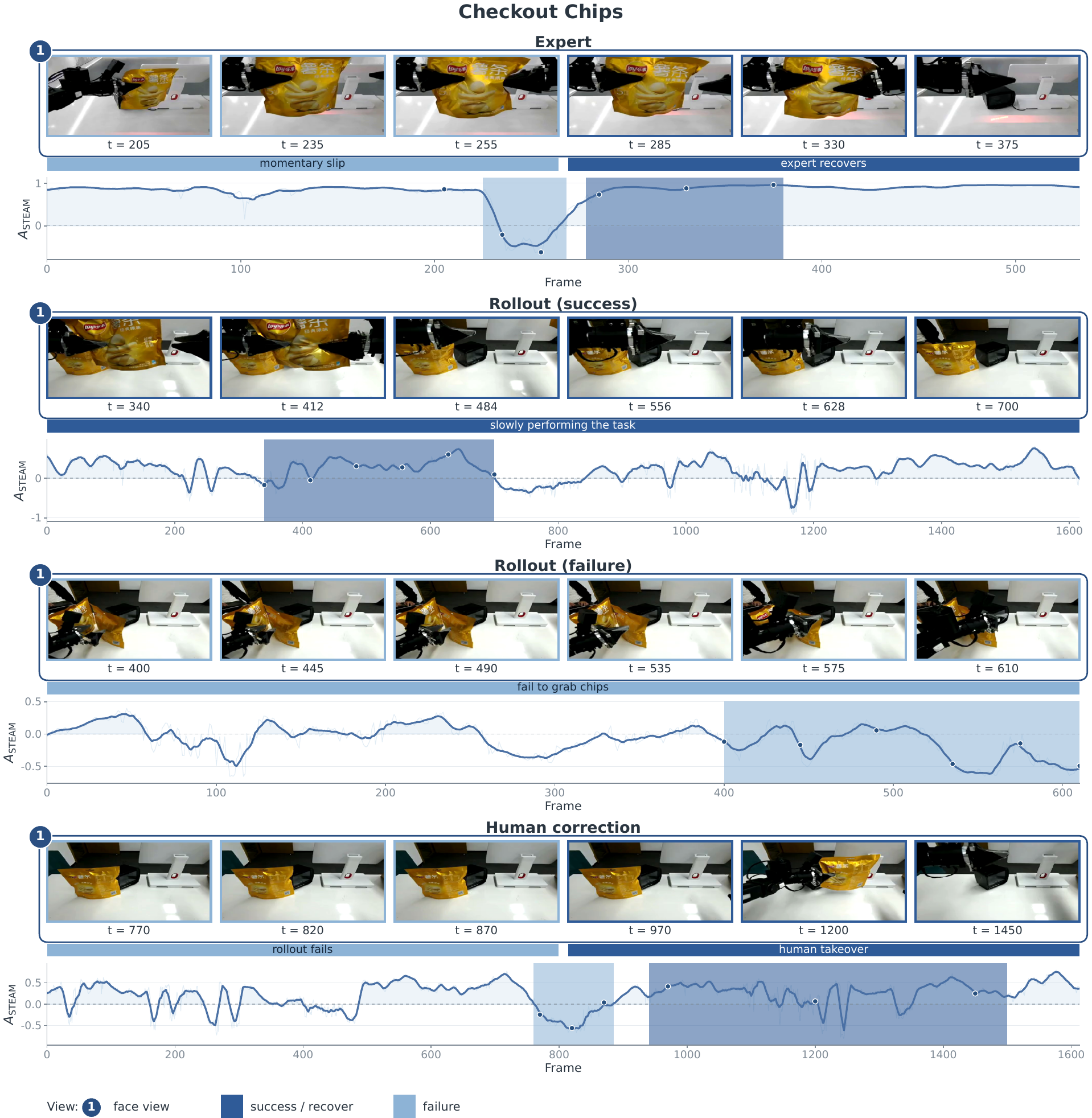}
  \caption{\textbf{Chip checkout advantage curves.} Expert, successful-rollout, failed-rollout, and human-correction episodes.}
  \label{fig:viz_chips_curve}
\end{figure}

\paragraph{Chip Checkout.}
For the \textbf{chip checkout} task (Figure~\ref{fig:viz_chips_curve}), the advantage curves accurately capture the complex dual-arm coordination and timing.
In the \textit{expert demonstration}, a brief grasping failure by the right arm during the handover phase (around frames 200–300) causes a temporary drop in the advantage; once a successful retry is executed, the advantage recovers to high values.
In the \textit{successful rollout}, the overall execution speed is visibly slower, particularly during the handover phase (indicated by the dark blue region), leading to a more fluctuating advantage curve.
The \textit{failed rollout} exhibits a persistent decline, with the advantage remaining extremely low after frame 400.
In the \textit{human-correction episode}, the policy stagnates around frame 800 and fails to grasp the second chip bag, causing the advantage to dip significantly into negative values. Following human takeover and corrective action, the task resumes progress, and the advantage recovers.

\begin{figure}[htbp]
  \centering
  \includegraphics[width=\linewidth]{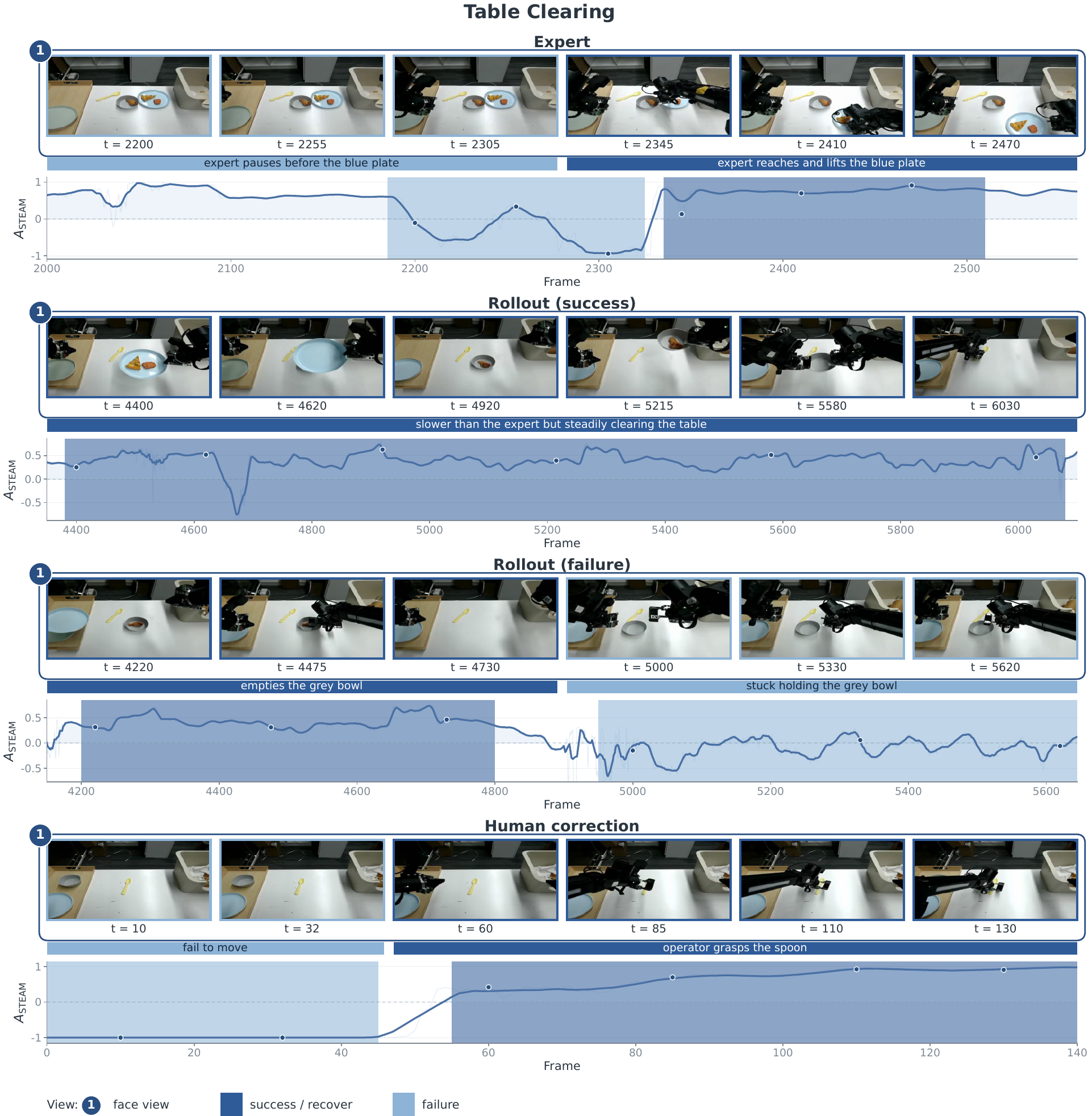}
  \caption{\textbf{Table clearing advantage curves.} Expert, successful-rollout, failed-rollout, and human-correction episodes.}
  \label{fig:viz_table_curve}
\end{figure}

\paragraph{Table Clearing.}
For the \textbf{table clearing} task (Figure~\ref{fig:viz_table_curve}), the advantage curves track pauses and stalls across the 21 stages.
In the \textit{expert demonstration}, after the green plate is put down, the right arm pauses out of view for about 7\,s (frames 2185--2325) before reaching for the blue plate in Stage 12; the advantage falls to around $-0.9$ during the pause and returns to $0.7$--$0.9$ once the arm reaches and lifts the plate.
In the \textit{successful rollout}, the policy clears the blue plate, the grey bowl, and the spoon at roughly half the expert pace, and the advantage stays at moderate positive values ($0.2$--$0.6$) apart from brief dips.
In the \textit{failed rollout}, the grey bowl is emptied and handed over to the left arm, but the left arm then holds it in the air and never puts it down (Stage 19); the advantage falls from $0.3$--$0.7$ to below zero after frame 4950 and stays low until the episode ends.
In the \textit{human-correction episode}, the robot fails to move at the start of Stage 20 and the advantage stays at $-1$; once the operator takes over and grasps the spoon, the advantage recovers to high values.

\begin{figure}[htbp]
  \centering
  \includegraphics[width=\linewidth]{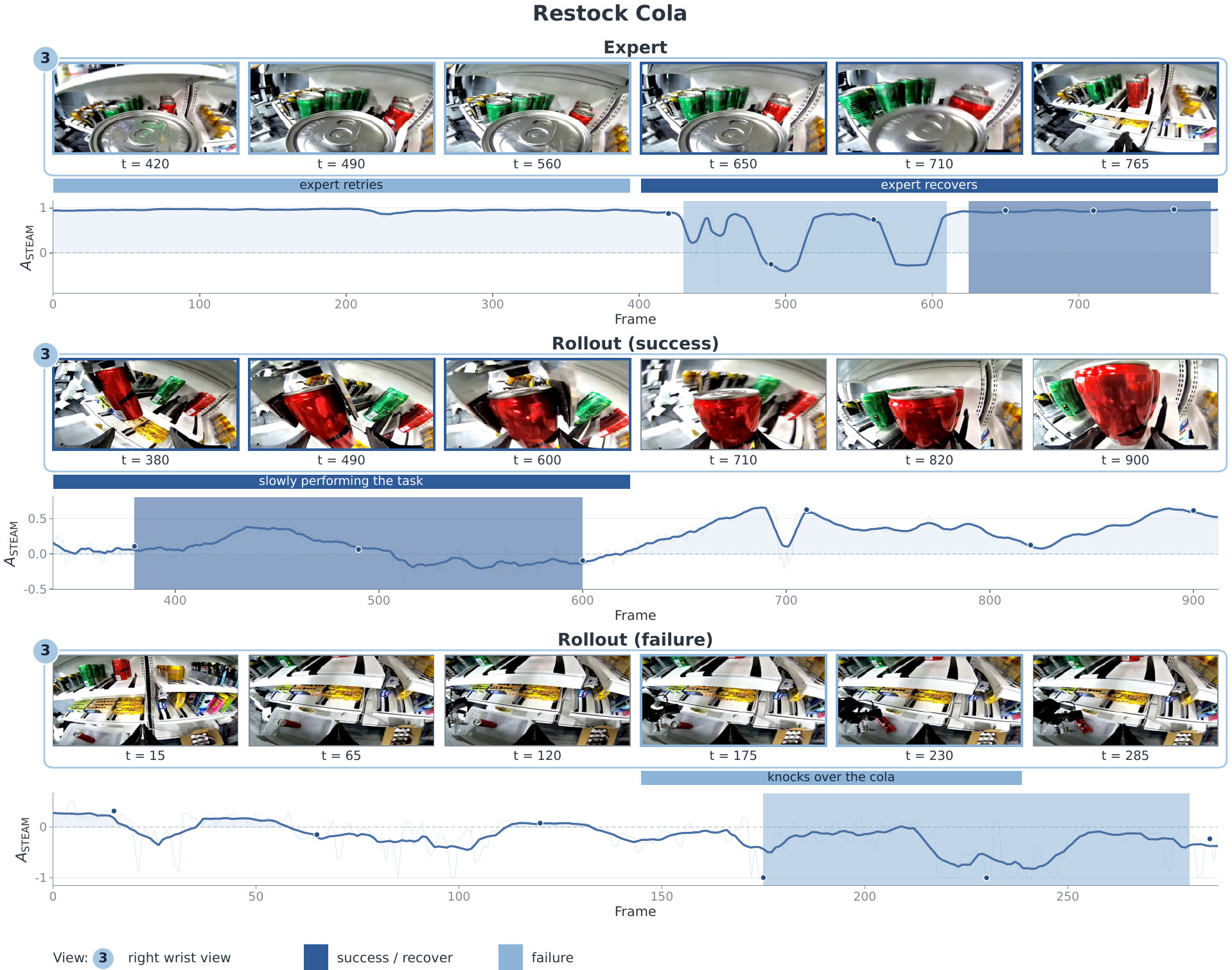}
  \caption{\textbf{Cola restocking advantage curves.} Expert, successful-rollout, and failed-rollout episodes.}
  \label{fig:viz_cola_curve}
\end{figure}

\paragraph{Cola Restocking.}
For the \textbf{cola restocking} task (Figure~\ref{fig:viz_cola_curve}), we observe similar temporal sensitivity to coordination and placement errors.
In the \textit{expert demonstration}, the robot fails to place the cola bottle onto the narrow shelf due to initial misalignment between frames 400 and 600, causing a temporary advantage drop. After a successful realignment and retry after frame 600, the advantage rises back to a high level.
In the \textit{successful rollout}, the right arm makes multiple consecutive attempts to grasp the cola during the handover phase, leading to slower progress and a corresponding decline in the advantage curve.
In the \textit{failed rollout}, the advantage drops and remains consistently low after frame 200, reflecting unrecoverable policy errors.

\subsection{Advantage Distributions Across Data Sources}
\label{app:adv_dist}

Section~\ref{sec:adv_eval} scores each episode by its cumulative advantage. This section reports the two distributions behind that number: how frame-level advantages are distributed within each data source, and how the per-episode sums separate.

\begin{figure*}[htb]
  \centering
  \begin{subfigure}[t]{0.24\linewidth}
    \centering
    \includegraphics[width=1\linewidth]{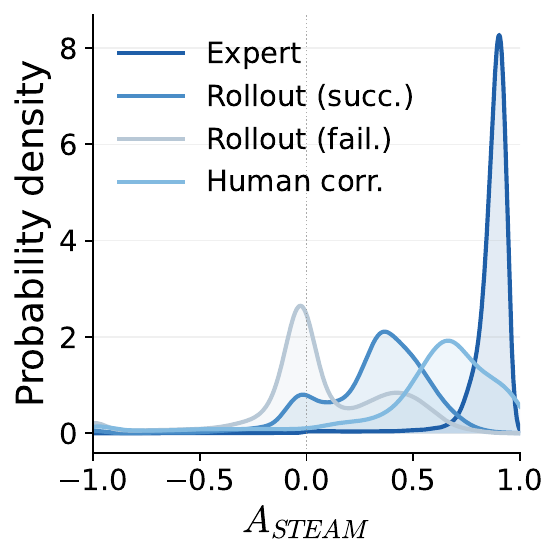}
    \caption{Towel Folding}
    \label{fig:density_towel}
  \end{subfigure}
  \hfill
  \begin{subfigure}[t]{0.24\linewidth}
    \centering
    \includegraphics[width=1\linewidth]{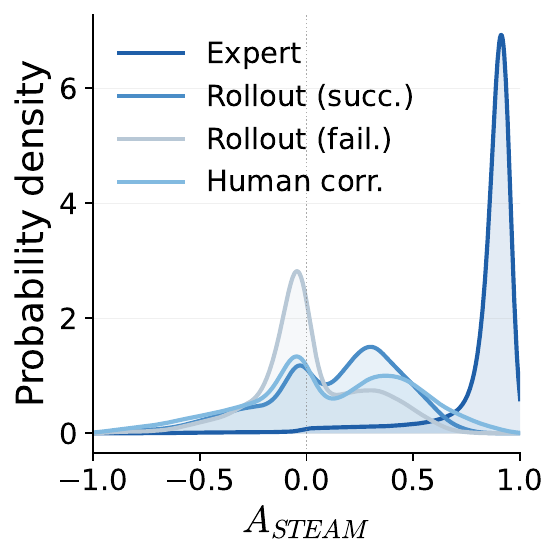}
    \caption{Chip Checkout}
    \label{fig:density_chips}
  \end{subfigure}
  \hfill
  \begin{subfigure}[t]{0.24\linewidth}
    \centering
    \includegraphics[width=1\linewidth]{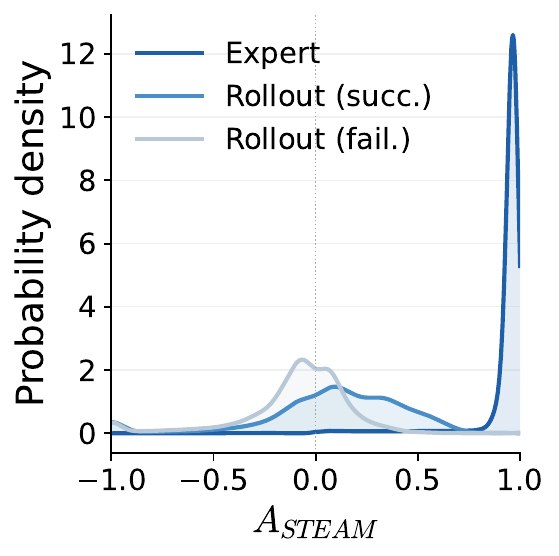}
    \caption{Cola Restocking}
    \label{fig:density_cola}
  \end{subfigure}
  \hfill
  \begin{subfigure}[t]{0.24\linewidth}
    \centering
    \includegraphics[width=1\linewidth]{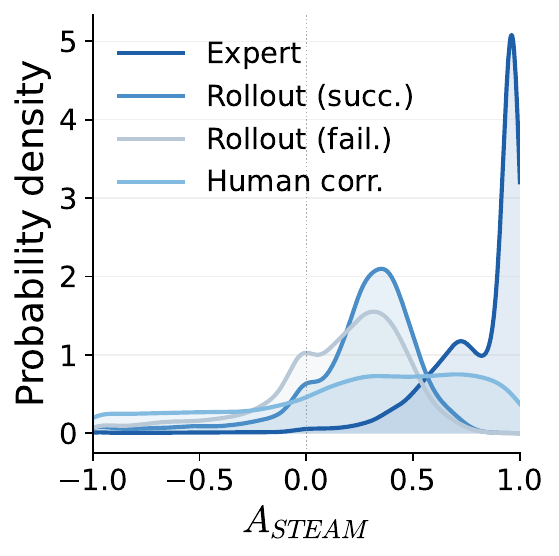}
    \caption{Table Clearing}
    \label{fig:density_table}
  \end{subfigure}
  \caption{Probability density of frame-level $A_{\mathrm{STEAM}}$ across data types.}
  \label{fig:density_all}
\end{figure*}

Figure~\ref{fig:density_all} presents the probability density of frame-level $A_{\mathrm{STEAM}}$ across different data sources.
Consistent trends are observed across all domains: expert demonstrations exhibit a strong concentration near $+1$, indicating that most frames correspond to high-quality execution that contributes directly to task completion.
Successful rollouts systematically shift toward lower positive values, reflecting slower or more cautious execution.
In contrast, failed rollouts place a prominent density peak around or below zero, which corresponds to the stagnation and lack of progress after a failure occurs.
Human-correction episodes exhibit a broader, more distributed profile, combining the lower positive advantages from slow autonomous phases with the higher advantages recovered immediately after human takeover.

\begin{figure*}[htb]
  \centering
  \begin{subfigure}[t]{0.24\linewidth}
    \centering
    \includegraphics[width=1\linewidth]{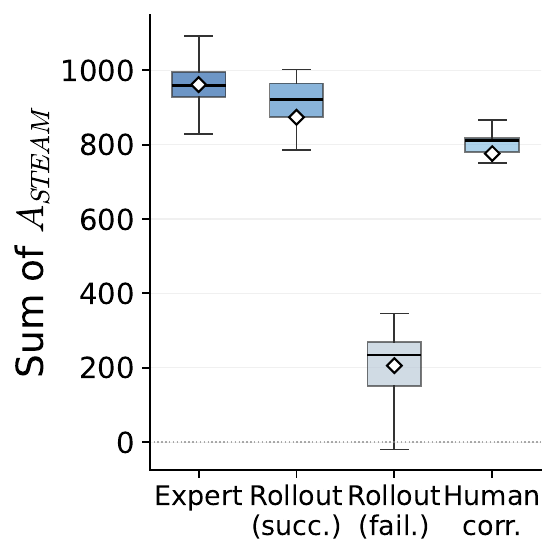}
    \caption{Towel Folding}
    \label{fig:sum_towel}
  \end{subfigure}
  \hfill
  \begin{subfigure}[t]{0.24\linewidth}
    \centering
    \includegraphics[width=1\linewidth]{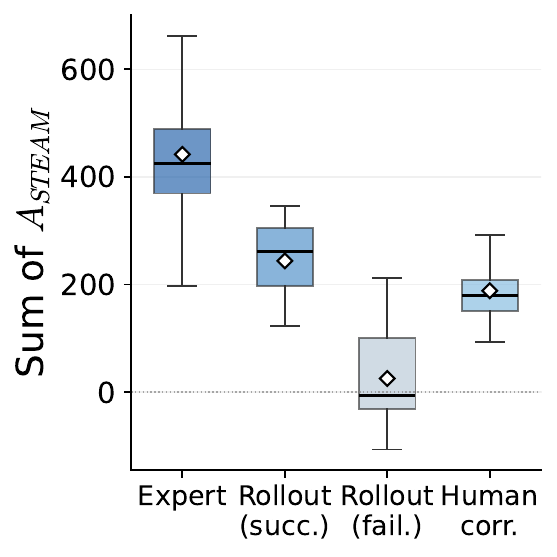}
    \caption{Chip Checkout}
    \label{fig:sum_chips}
  \end{subfigure}
  \hfill
  \begin{subfigure}[t]{0.24\linewidth}
    \centering
    \includegraphics[width=1\linewidth]{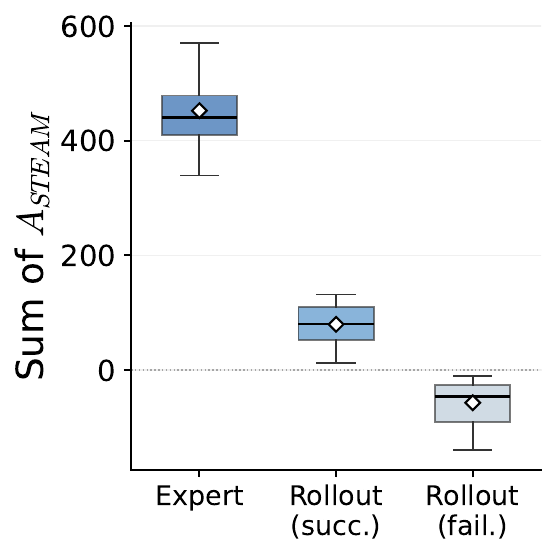}
    \caption{Cola Restocking}
    \label{fig:sum_cola}
  \end{subfigure}
  \hfill
  \begin{subfigure}[t]{0.24\linewidth}
    \centering
    \includegraphics[width=1\linewidth]{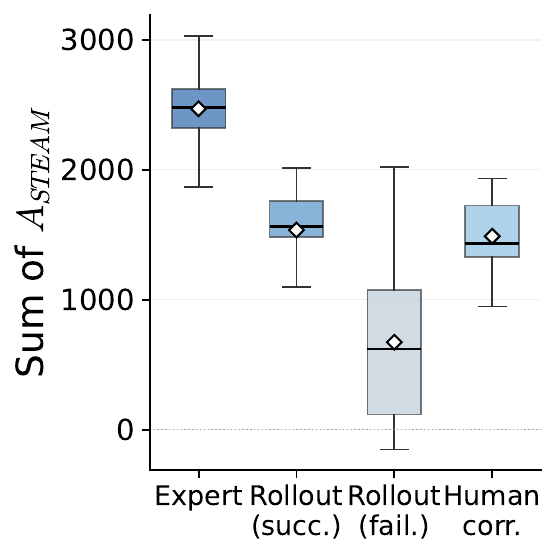}
    \caption{Table Clearing}
    \label{fig:sum_table}
  \end{subfigure}
  \caption{Sum of frame-level $A_{\mathrm{STEAM}}$ over each episode across data types.}
  \label{fig:sum_all}
\end{figure*}

Figure~\ref{fig:sum_all} reports the episode-wise sum of frame-level advantages, the quantity clustered in Table~\ref{tab:adv_quality}.
While successful episodes yield comparable cumulative advantages, failed rollouts are clearly separated by significantly lower sums.
Notably, cumulative advantages are higher for towel folding and chip checkout than for cola restocking, particularly among failures.
This discrepancy stems from their failure modes: towel and chip failures typically occur in middle or late stages, allowing positive advantages to accumulate early on, whereas cola failures occur during the initial grasp, preventing any advantage accumulation.
The absolute scale of the sum therefore carries task-specific information and is not comparable across tasks, which is why Table~\ref{tab:adv_quality} clusters episodes within a single task.

\subsection{Implementation Details: Advantage Modeling and CFGRL Policy Extraction}
\label{app:cfgrl}

This section expands the policy-training step sketched in Equation~\plaineqref{eq:label}: how the frame-level advantages learned by STEAM enter the classifier-free guidance reinforcement learning framework, and how training and inference proceed once they do.

\paragraph{CFGRL Training.}
Once the STEAM advantage predictors are trained on the expert dataset, we compute the advantage score $A_{\mathrm{STEAM}}(f_{k,i}; \Theta)$ for each frame $f_{k,i} \in \Dexp \cup \Dnexp$ using a fixed lookahead $H$.
Here, $\Dexp$ represents the expert demonstration dataset, and $\Dnexp$ denotes the non-expert dataset, comprising both autonomous rollouts and human corrections.
We then map these continuous advantages into binary optimality labels $o \in \{0, 1\}$ by thresholding each data pool separately:
\begin{equation}
  o =
  \begin{cases}
    \mathbf{1}\!\left[A_{\mathrm{STEAM}}(f_{k,i}; \Theta) \ge \delta^{\mathrm{exp}}\right],
    & f_{k,i} \in \Dexp,\\[2pt]
    \mathbf{1}\!\left[A_{\mathrm{STEAM}}(f_{k,i}; \Theta) \ge \delta^{\mathrm{non\text{-}exp}}\right],
    & f_{k,i} \in \Dnexp,
  \end{cases}
  \label{eq:cfg-label}
\end{equation}
where $\delta^{\mathrm{exp}}$ and $\delta^{\mathrm{non\text{-}exp}}$ label the top-$q^{\mathrm{exp}}$ and top-$q^{\mathrm{non\text{-}exp}}$ fractions of $\Dexp$ and $\Dnexp$ as optimal.

These binary optimality labels condition a flow-matching policy $v_\phi(a_t, t, s, o)$ parameterizing a velocity field. During training, we construct a noisy action interpolant $a_t$ at timestep $t$ as:
\begin{equation}
  a_t = (1-t)a_0 + t a,
\end{equation}
where $a$ is the target action chunk from the dataset, $a_0 \sim \mathcal{N}(0, I)$ is standard Gaussian noise, and $t$ is sampled from a uniform distribution over $[0, 1]$, denoted as $t \sim \mathcal{U}[0,1]$.

To support classifier-free guidance, we randomly drop the conditioning label $o$ with a dropout probability $p_{\mathrm{drop}}$. The conditioned label input $\tilde{o}$ is defined as $\tilde{o} = \varnothing$ (representing the unconditioned state) with probability $p_{\mathrm{drop}}$, and $\tilde{o} = o$ otherwise. The resulting flow matching objective is formulated as:
\begin{equation}
  \mathcal{L}_\pi(\phi) =
  \mathbb{E}_{(s,a,o)\sim \Dexp \cup \Dnexp,\; t \sim \mathcal{U}[0,1],\; a_0 \sim \mathcal{N}(0, I)}
  \left[
    \left\| v_\phi(a_t, t, s, \tilde{o}) - (a - a_0) \right\|^2
  \right],
  \label{eq:cfgloss}
\end{equation}
where $s$ denotes the policy's observation input of the current frame $f_{k,i}$.

\paragraph{CFGRL Inference.}
During deployment, we generate guided actions by integrating the ODE defined by the velocity field. Starting from pure Gaussian noise $a_0 \sim \mathcal{N}(0, I)$ at $t=0$, we perform $T$ fixed-size Euler integration steps with step size $\Delta t = 1/T$:
\begin{equation}
  a_{t+\Delta t} = a_t + \Delta t \cdot v_{\mathrm{cfg}}(a_t, t, s),
\end{equation}
where the guided velocity field $v_{\mathrm{cfg}}(a_t, t, s)$ is computed by extrapolating towards the high-optimality label ($o=1$):
\begin{equation}
  v_{\mathrm{cfg}}(a_t, t, s)
  =
  v_\phi(a_t, t, s, \varnothing)
  +
  w\Bigl[
    v_\phi(a_t, t, s, o \!=\! 1)
    -
    v_\phi(a_t, t, s, \varnothing)
  \Bigr].
  \label{eq:cfg}
\end{equation}
Here, $w \ge 1$ is the guidance scale that controls the strength of the optimality bias. Setting $w > 1$ actively steers the generated actions toward trajectories classified as highly progress-advancing by STEAM. After $T$ integration steps, the resulting endpoint $a_1$ is sent to the robot controller as the target action chunk.

\paragraph{Pseudocode.}
We provide the formal algorithmic procedures for our framework. Specifically, Algorithm~\ref{alg:train} details the self-supervised training process for the STEAM advantage predictor ensemble. Algorithm~\ref{alg:cfgrl} describes the policy learning pipeline, showing how learned advantages are translated into binary optimality labels to guide a classifier-free flow-matching policy.
\label{app:pseudocode}

\begin{algorithm}[!htbp]
\caption{Training STEAM}
\label{alg:train}
\begin{algorithmic}[1]
\Input Expert dataset $\Dexp$, ensemble size $M$, bin count $N$,
       max offset $k_{\max}$, reference length $L_{\max}$,
       offset bound $\tilde\Delta_{\max}$
\Output Trained ensemble $\Theta = \{\theta_m\}_{m=1}^{M}$
\For{$m = 1$ \textbf{to} $M$}
  \State Initialize $\theta_m$ with an independent random seed
  \Repeat
    \State Sample expert episode $\tau_k = (f_{k,1}, \ldots, f_{k,L_{\tau_k}})$ with instruction $\ell$ from $\Dexp$
    \State Sample offset $\Delta \sim \mathrm{Uniform}\{\pm 1, \ldots, \pm k_{\max}\}$
    \State Sample frame index $i \sim \mathrm{Uniform}\{\, i' : 1 \le i' \le L_{\tau_k},\; 1 \le i' + \Delta \le L_{\tau_k} \,\}$;\; set $j \leftarrow i + \Delta$
    \State $\tilde\Delta_{\tau_k}(i, j) \leftarrow (j - i) \cdot L_{\max} / L_{\tau_k}$
    \hfill \Comment{Eq.~\plaineqref{eq:ktilde}}
    \State $y \leftarrow \tilde\Delta^{\mathcal{B}}_{\tau_k}(i, j)$
    \hfill \Comment{one-hot bin of $\mathrm{clip}(\tilde\Delta_{\tau_k}(i, j),\, \pm\tilde\Delta_{\max})$}
    \State $\mathcal{L} \leftarrow H\!\bigl(y,\; p_{\theta_m}(\Delta \mid f_{k,i}, f_{k,j}, \ell)\bigr)$
    \hfill \Comment{Eq.~\plaineqref{eq:loss}}
    \State Update $\theta_m$ via gradient step on $\mathcal{L}$
  \Until{convergence}
\EndFor
\State \Return $\Theta = \{\theta_m\}_{m=1}^{M}$
\end{algorithmic}
\end{algorithm}

\begin{algorithm}[!htbp]
\caption{CFGRL Policy Learning with STEAM}
\label{alg:cfgrl}
\begin{algorithmic}[1]
\Input Trained STEAM ensemble $\Theta = \{\theta_m\}_{m=1}^{M}$, expert dataset
       $\Dexp$, non-expert dataset $\Dnexp$, policy lookahead
       $H$, top fractions $q^{\mathrm{exp}}$ and $q^{\mathrm{non\text{-}exp}}$, conditioning
       dropout $p_{\mathrm{drop}}$
\Output Trained flow-matching policy $v_\phi$
\Statex \textit{\textbf{Stage 1: label data with STEAM}}
\For{each $(s_t, a_t, s_{t+H}) \in \Dexp \cup \Dnexp$}
  \State $\Asteam(s_t, s_{t+H}) \leftarrow
         \displaystyle\min_{m=1,\ldots,M} A(s_t, s_{t+H};\, \theta_m)$
  \hfill \Comment{Eqs.~\plaineqref{eq:advantage},~\plaineqref{eq:ensemble}}
\EndFor
\State $\delta^{\mathrm{exp}} \leftarrow \text{top-}q^{\mathrm{exp}}\text{ threshold of }
       \{\Asteam(s_t,s_{t+H})\}\text{ over }\Dexp$
\State $\delta^{\mathrm{non\text{-}exp}} \leftarrow \text{top-}q^{\mathrm{non\text{-}exp}}\text{ threshold of }
       \{\Asteam(s_t,s_{t+H})\}\text{ over }\Dnexp$
\For{each $(s_t, a_t, s_{t+H}) \in \Dexp \cup \Dnexp$}
  \If{$(s_t, a_t, s_{t+H}) \in \Dexp$}
    \State $o_t \leftarrow \mathbf{1}\!\left[\Asteam(s_t, s_{t+H}) \ge \delta^{\mathrm{exp}}\right]$
  \Else
    \State $o_t \leftarrow \mathbf{1}\!\left[\Asteam(s_t, s_{t+H}) \ge \delta^{\mathrm{non\text{-}exp}}\right]$
  \EndIf
\EndFor
\State $\mathcal{D}_o \leftarrow \{(s_t, a_t, o_t)\}$
\hfill \Comment{Eq.~\plaineqref{eq:cfg-label}}
\Statex \textit{\textbf{Stage 2: classifier-free guidance policy training}}
\State Initialize flow-matching policy parameters $\phi$
\Repeat
  \State Sample $(s, a, o) \sim \mathcal{D}_o$
  \State Sample flow time $t \sim \mathcal{U}[0,1]$ and noise
         $a_0 \sim \mathcal{N}(0, I)$
  \State $a_t \leftarrow (1-t)\,a_0 + t\,a$
  \State $\tilde{o} \leftarrow \varnothing$ with probability
         $p_{\mathrm{drop}}$, otherwise $\tilde{o} \leftarrow o$
  \State $\mathcal{L}_\pi \leftarrow
         \bigl\| v_\phi(a_t, t, s, \tilde{o}) - (a - a_0) \bigr\|^2$
  \hfill \Comment{Eq.~\plaineqref{eq:cfgloss}}
  \State Update $\phi$ via gradient step on $\mathcal{L}_\pi$
\Until{convergence}
\State \Return $v_\phi$
\end{algorithmic}
\end{algorithm}

\paragraph{Hyperparameters.}
\label{app:hyperparams}
Table~\ref{tab:hyperparams} lists the hyperparameters of the STEAM advantage predictor and the CFGRL policy training, which are shared by all four tasks.
The lookahead $H$ is the one setting that is worth calling out: it fixes the temporal span over which progress is measured, and it has to be long enough for the scene to change visibly. Our default $H=32$ spans $1.6$\,s at the $20$\,fps the data is recorded at; halving it to $H=16$ ($0.8$\,s) lowers the towel-folding success rate from $92.3\%$ to $80.0\%$, because over such a short interval the visual difference between two frames is too subtle to signify progress.

\begin{table}[!htbp]
  \caption{Hyperparameters for STEAM advantage prediction and CFGRL policy optimization, shared by all four tasks.}
  \label{tab:hyperparams}
  \centering
  \small
  \begin{tabular}{lc}
    \toprule
    \textbf{Hyperparameter} & \textbf{Value} \\
    \midrule
    Maximum offset            & 32 \\
    Number of bins $N$        & 32 \\
    Ensemble size $M$         & 3 \\
    Advantage lookahead $H$   & 32 \\
    Top fraction $q^{\mathrm{exp}}$ over $\Dexp$ & 0.8 (top 80\%) \\
    Top fraction $q^{\mathrm{non\text{-}exp}}$ over $\Dnexp$ & 0.3 (top 30\%) \\
    Conditioning dropout $p_{\mathrm{drop}}$ & 0.1 \\
    Guidance scale $w$        & 2.5 \\
    Learning rate             & $5\times10^{-5}$ \\
    Training steps            & 30,000 \\
    Batch size                & 512 \\
    \bottomrule
  \end{tabular}
\end{table}

\FloatBarrier
\subsection{Continuous Advantage Weighting as an Alternative to CFGRL}
\label{app:awr}

CFGRL turns $A_{\mathrm{STEAM}}$ into a binary optimality label and conditions the policy on it (Eq.~\plaineqref{eq:cfg-label}). A natural alternative is to keep the advantage continuous and let it reweight the imitation loss directly, as in advantage-weighted regression~\citep{peng2019awr}, where each transition is weighted by $\exp(A/\beta)$. This is a strictly finer-grained use of the same signal, so it is worth asking whether the binarization in CFGRL throws information away.

Table~\ref{tab:awr} answers this on towel folding, with the advantage model, the dataset, and the policy backbone held fixed and only the policy-extraction rule changed. AWR recovers a large part of the gap over behavior cloning, from $33.3\%$ to $73.3\%$, which confirms that the STEAM advantage is informative under either extraction rule. But it remains $19$ points below CFGRL.

We attribute the gap to the bounded range of the advantage. Since $A_{\mathrm{STEAM}} \in [-1, 1]$ by construction, $\exp(A/\beta)$ spans a narrow band of weights, and continuous weighting is not sharp enough to filter failed transitions: a stalled or regressing frame is down-weighted, but it still contributes a non-trivial gradient, and in datasets where such frames are a large fraction of the non-expert data their aggregate pull is substantial. Quantile thresholding instead removes those frames from the optimal mode entirely, and the guidance scale $w$ then controls at inference time how strongly the policy is pushed toward that mode, decoupling the strength of the optimality bias from the scale of the advantage. This is consistent with the finding in RECAP~\citep{amin2025recap} that advantage-conditioned policy extraction outperforms advantage-weighted regression.

\begin{table}[ht]
  \caption{\textbf{Policy extraction rule on towel folding.} Success rate (\%) with the STEAM advantage held fixed and only the extraction rule changed.}
  \label{tab:awr}
  \centering
  \small
  \begin{tabular}{lccc}
    \toprule
    & \textbf{BC} & \textbf{STEAM (AWR)} & \textbf{STEAM (CFGRL, ours)} \\
    \midrule
    Success rate (\%) & 33.3 & 73.3 & \textbf{92.3} \\
    \bottomrule
  \end{tabular}
\end{table}

\FloatBarrier
\subsection{When Expert-Only Filtering Does Not Help: The Pick-and-Place Case}
\label{app:pnp_limitation}

STEAM improves policy performance by filtering out low-quality, non-progressing frames and emphasizing the most task-advancing segments. The size of that benefit depends on what there is to filter. When the available data are both scarce and already of high quality, filtering removes samples that a supervised policy could still have learned from, effectively shrinking an already small training set. We study this regime on the \textbf{pick-and-place} task, a short-horizon task on a single Franka arm in which the gripper relocates objects between two plates. Expert trajectories here are highly consistent, and the dataset contains 50 expert demonstrations and 594 autonomous rollouts. The task has 2 stages:
\begin{itemize}
    \item \textbf{Stage 1 (Grasp):} The gripper descends to grasp a target object from the left plate.
    \item \textbf{Stage 2 (Place):} The arm transfers and places the object onto the designated position on the right plate.
\end{itemize}
Because the episodes are much shorter, we halve the maximum offset, the bin count $N$, and the lookahead $H$ to 16 on this task; all other hyperparameters follow Table~\ref{tab:hyperparams}.

\begin{figure}[htbp]
  \centering
  \includegraphics[width=0.7\linewidth]{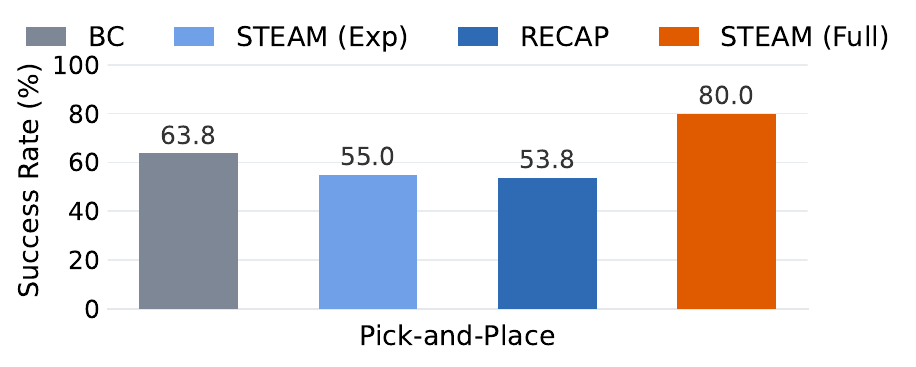}
  \caption{\textbf{Policy performance on the pick-and-place task.} $\text{STEAM (Exp)}$ filters expert demonstrations only, whereas $\text{STEAM (Full)}$ additionally uses autonomous rollouts. BC vs.\ $\text{STEAM (Exp)}$ shows that filtering the already-clean expert data alone does not help, while RECAP vs.\ $\text{STEAM (Full)}$ shows that STEAM captures the useful signal in rollout data more effectively than RECAP.}
  \label{fig:pnp_perf}
\end{figure}

Figure~\ref{fig:pnp_perf} reports the policy performance, and the four-way comparison isolates two effects. First, STEAM trained on expert demonstrations only ($\text{STEAM (Exp)}$) slightly underperforms the already-strong BC baseline ($63.8\%$). The expert set is small (50 trajectories) and already high-quality, so filtering it only removes useful supervision and lowers performance. This near-optimality is directly visible in Figure~\ref{fig:pnp_density}, where the frame-level $A_{\mathrm{STEAM}}$ of expert demonstrations concentrates sharply near $+1$, leaving almost no low-quality mass to remove. Second, once autonomous rollouts are added ($\text{STEAM (Full)}$), STEAM selects the high-quality frames from this rollout data and raises the success rate to $80\%$, surpassing both BC and RECAP.

The limitation is therefore specific, and narrower than the section title suggests on its own: what does not help on this task is filtering an expert set that is already clean and small. Given the full data it is applied to everywhere else in this paper, expert demonstrations together with autonomous rollouts and human corrections, STEAM still outperforms BC on pick-and-place, by $16$ points. The regime to be careful about is not short-horizon tasks in general but datasets that offer nothing to prune.

\begin{figure}[htbp]
  \centering
  \includegraphics[width=\linewidth]{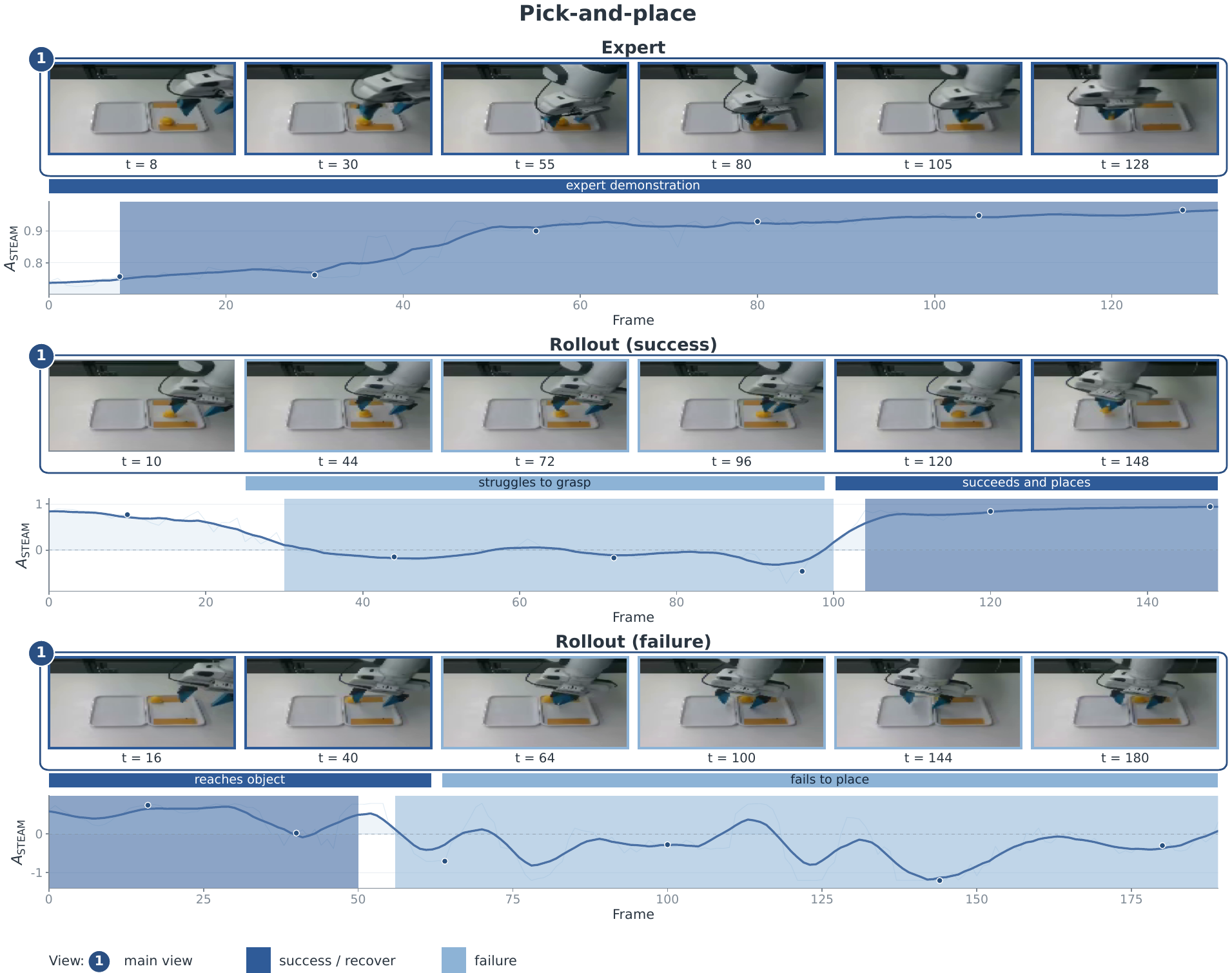}
  \caption{\textbf{Pick-and-place advantage curves.} Expert, successful-rollout, and failed-rollout episodes.}
  \label{fig:viz_pnp_curve}
\end{figure}

\begin{figure}[htbp]
  \centering
  \begin{minipage}[b]{0.56\linewidth}
    \centering
    \includegraphics[height=3.4cm]{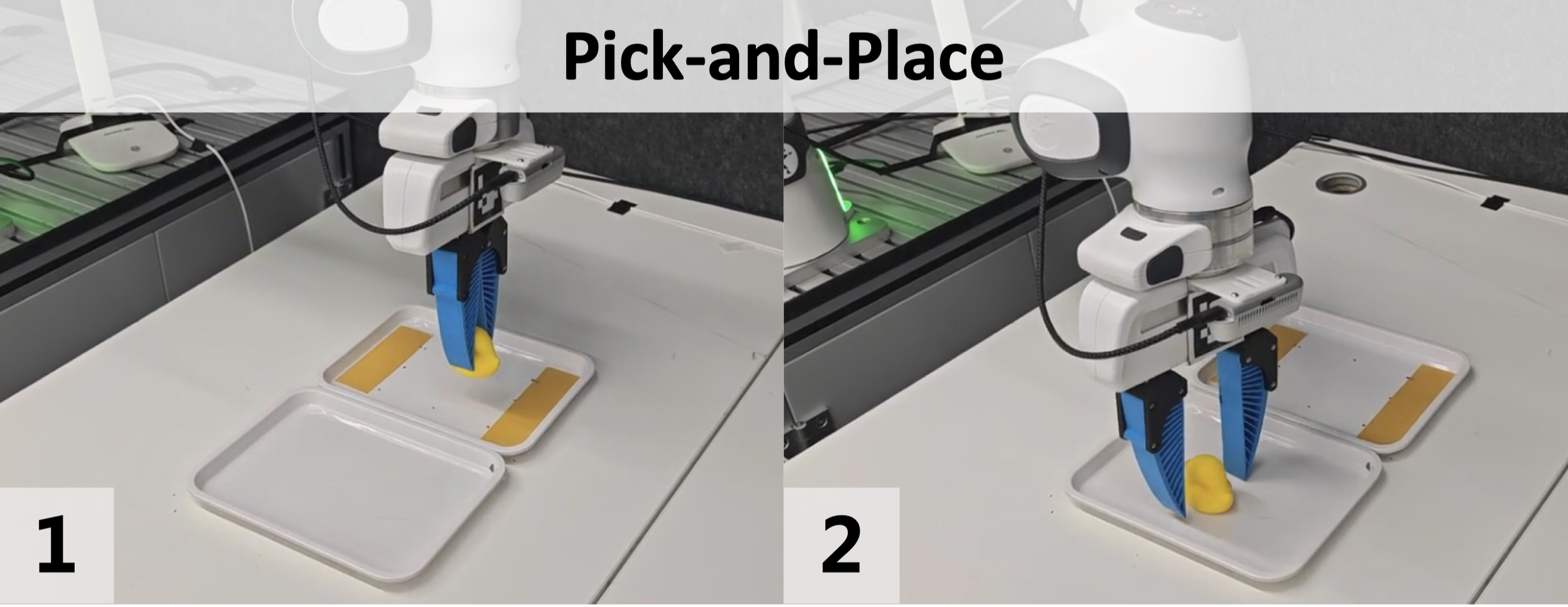}
    \\[0.4em]
    {\small (a) Task setup}
  \end{minipage}\hfill
  \begin{minipage}[b]{0.40\linewidth}
    \centering
    \includegraphics[height=3.4cm]{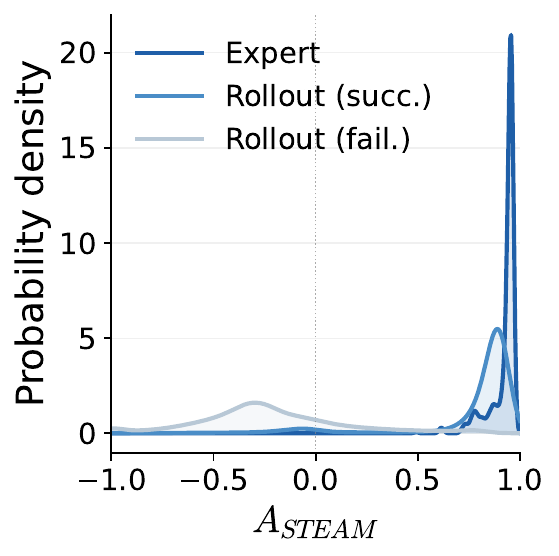}
    \\[0.4em]
    {\small (b) Advantage density}
  \end{minipage}
  \caption{\textbf{The pick-and-place task and its frame-level advantage density.} \textbf{(a)} Pick-and-place is a short, two-stage task (grasp the object, then place it), so expert trajectories are highly consistent. \textbf{(b)} The frame-level $A_{\mathrm{STEAM}}$ of expert demonstrations concentrates sharply near $+1$, indicating that trajectories are already near-optimal and leave little low-quality mass for filtering to prune. This is why $\text{STEAM (Exp)}$ slightly underperforms BC in this scarce, already-clean regime.}
  \label{fig:pnp_density}
\end{figure}

Figure~\ref{fig:viz_pnp_curve} shows the frame-level advantage curves on this task for completeness. Even at this much shorter horizon the signal remains informative.
The \textit{expert demonstration} shows smooth, proficient execution, maintaining a high advantage near $+1$ throughout.
In the \textit{successful rollout}, the robot makes multiple unsuccessful attempts to grasp the target object (a toy duck) between frames 40 and 100, pushing the advantage below zero. After a successful grasp after frame 100, the advantage recovers to nearly $+1$.
In the \textit{failed rollout}, the advantage is initially high (frames 0--50) as the gripper approaches the duck, but plunges below zero once the robot repeatedly fails to grasp the object.

\end{document}